\documentclass{article}

\usepackage{amsmath, amssymb, amsthm}
\usepackage{microtype}
\usepackage{graphicx}
\usepackage{subfigure}
\usepackage{caption}

\usepackage{bm}
\usepackage{booktabs} 
\usepackage[usenames,dvipsnames]{color}
\usepackage{wrapfig}
\usepackage{multirow}
\newtheorem{prop}{Proposition}
\usepackage{hyperref}

\usepackage{algorithm}
\usepackage{algpseudocode}

\usepackage[accepted]{mlsys2025}

\mlsystitlerunning{A Continuous Relaxation of Quantization-Aware Training}

\begin{document}

\twocolumn[
\mlsystitle{A Continuous Relaxation of Quantization-Aware Training}




\begin{mlsysauthorlist}
\mlsysauthor{Kartikeya Bhardwaj}{qair}
\mlsysauthor{Mohammed Almousa}{qair}
\mlsysauthor{Chaitanya Mullapudi}{qair}
\mlsysauthor{Ananth Balasubramanian}{qair}
\mlsysauthor{Nithin Ramesan}{qair}
\mlsysauthor{Tuc Nguyen}{qair}
\mlsysauthor{Harris Teague}{qair}
\end{mlsysauthorlist}

\mlsysaffiliation{qair}{Qualcomm AI Research. Qualcomm AI Research is an initiative of Qualcomm Technologies, Inc}

\mlsyscorrespondingauthor{Kartikeya Bhardwaj}{kbhardwa@qti.qualcomm.com}

\mlsyskeywords{Machine Learning, MLSys}

\vskip 0.3in

\vspace{-2mm}
\begin{abstract}
Quantization is a critical piece of the deployment pipeline in the era of large generative models. Low-bit quantization, particularly 2-bits or sub-2-bits, is a notoriously difficult problem, particularly when using a channelwise quantization scheme which has been shown as a leading candidate for hardware-efficiency. The key challenges with low-bit quantization range from limited quantization levels to saddle points or flat regions in the loss surface. In this paper, we look at the problem of flat quantization loss surfaces through a lens of dynamically changing bitwidth. Unlike existing techniques which either change the bit-precision abruptly or keep it fixed, we propose to gradually alter the bitwidth as the model trains. To this end, we propose RelaxQAT, a continuous relaxation of the QAT process. We also provide simple theoretical and empirical insights behind why RelaxQAT helps the QAT performance. Our extensive experiments demonstrate up to $\bm{6.3\%}$ average accuracy gains on vision and up to $\bm{7.7\%}$ better average accuracy on natural language tasks, compared to the leading QAT methods.
\end{abstract}
]



\printAffiliationsAndNotice{}

\section{Introduction}
Quantization has become the de facto strategy for deploying Large Language Models (LLMs) and Vision-Language models (VLMs) on resource-constrained systems such as phones and laptops~\cite{nagel2021white, gong2025survey}. To address the substantial memory footprint, power consumption, and cycle cost of inference of generative models, quantization reduces the precision of weights and activations from a higher bitwidth to a significantly lower bitwidth. This reduction in precision introduces quantization errors which result in noisy model outputs and an associated accuracy degradation in the real-world applications~\cite{song2026target}. The prevailing paradigm that current hardware supports is quantization bitwidths in the range of 2 to 4 bits. The lower the bits, the worse the quantization error, albeit with a significantly better energy-efficiency. Although many advanced  techniques exist that address 4-bit precision~\cite{lin2024awq, liu2025spinquant, sun2020ultra, frantar2022gptq, ashkboos2024quarot}, even lower bitwidth, such as 2-bit quantization, remains a critical challenge~\cite{paretoq, winq, upq}. Hence, investigating newer quantization techniques to address 2-bit precision for weights is of paramount importance.

For such low precisions, training-free methods known as Post-Training Quantization (PTQ) have \textit{not} been effective at recovering the quantization error~\cite{paretoq, yi2026jacquant}. Consequently, Quantization-Aware Training (QAT) techniques have emerged as the standard methods to convert a pretrained floating point model to a 2-bit weight model~\cite{song2026target}. Various QAT methods target different challenges: LSQ~\cite{lsq} proposes a learned scale as well as improvements to the gradient computation to improve the QAT convergence. ParetoQ~\cite{paretoq} proposes a new symmetric integer grid for 1.58- and 2-bit models. WinQ~\cite{winq} addresses the presence of saddle points or flat regions in the quantized loss surface. Progressive quantization methods such as UPQ~\cite{upq} focus on converting a model from a floating point precision to 2-bits in multiple stages, e.g., going from floating point model to 4-bits first, and then eventually quantizing the model to 2-bits. 

Most QAT techniques in the literature result in an abrupt change in bitwidth, either from floating point directly to 2-bits~\cite{llmqat, chen2025efficientqat, paretoq, winq, yi2026jacquant}, or as in the case of UPQ~\cite{upq}, from floating point to 4-bits first and then suddenly to 2-bits. Once the desired bitwidth is reached, the model is trained with no further changes in precision. It is known that low-bit quantization loss surfaces usually consist of highly flat surfaces~\cite{winq} due to the nature of quantization itself. Hence, when we optimize on a fixed bitwidth, the optimization may get stuck in those flat portions of the loss surface. Indeed, increasing the bit-precision of the weights can make the loss surface less flat and can make it easier for the optimization to escape saddle points. To this end, we hypothesize whether a dynamic bitwidth during the QAT can be a solution to this problem. Importantly, since abrupt changes in bitwidth can destabilize the training, we need a solution that makes gradual changes to the bitwidth as the training progresses.

Therefore, we ask the following \textbf{key questions} in this paper: \textit{During QAT,  can we dynamically relax the bitwidth of a model in a continuous fashion?} That is, instead of discrete (abrupt) integer changes in the number of bits, can we change the bitwidth in a continuous (floating point) domain? If yes, how do we implement a floating point bitwidth and what does it even mean?

To answer the above questions, we propose \textit{RelaxQAT}, a continuous relaxation of the QAT problem. The key idea behind RelaxQAT is to create a schedule for the bitwidth and change the underlying QAT equations, so that the \textit{extreme values} in the model weight distributions become floating points which slowly turn into integers during the training. Importantly, the dynamically changing bitwidth also dynamically changes the number of quantization bins, which helps the model escape saddle points and reach a better training loss. With extensive experiments, we demonstrate that our method achieves significantly better benchmark accuracies on a variety of vision-language and natural language tasks. Of note, our work focuses only on channelwise quantization since it remains one of the most hardware-efficient quantization schemes, and is highly relevant to deploying large generative models on constrained hardware such as Neural Processing Units (NPUs).

Overall, we make the following \textbf{key contributions}:
\begin{enumerate}
    \item We propose RelaxQAT, a continuous relaxation of the QAT. To our knowledge, we are the first to dynamically vary the bitwidth in a floating point domain and the resulting number of quantization bins during the QAT.
    \item We provide simple theoretical and empirical insights behind why RelaxQAT helps the QAT performance.
    \item For the challenging case of 2-bit weight only quantization, we demonstrate that RelaxQAT achieves the state-of-the-art accuracies on a variety of vision-language and natural language benchmarks. We show an improvement in average accuracy of up to $\bm{6.3\%}$ on vision benchmarks compared to the leading QAT baselines. On natural language tasks, we show up to $\bm{7.7\%}$ better average accuracy. In total, we conduct experiments on five models of various  sizes (0.6B to 4B), spanning two VLMs, three LLMs, and multiple model (Gemma and Qwen) families. We consider five leading QAT baselines and each model is trained on 10B tokens.
    \item Finally, we also show a proof-of-concept that RelaxQAT can be used for other bitwidths such as ternary (1.58-bit) quantization. We show up to $\bm{4.2\%}$ accuracy gains for this quantization scheme.
\end{enumerate}

\section{Related Work}\label{sec::rel}
We review briefly important quantization methods from the literature to set the stage for our proposed RelaxQAT algorithm. One of the early work in low-bit quantization was the Learned Step Quantization, introduced by \cite{lsq}. LSQ's key innovation was to treat the quantization step size itself as a learnable parameter .\cite{choi2018pact}, proposed learning activation clipping thresholds during training while \cite{bhalgat2020lsq+} proposed learning both scale and offset.

Building on LSQ, \cite{paretoq}  proposed ParetoQ which introduced Stretched Elastic Quantizer (SEQ) for ternary and 2-bit quantization. It is shown that although LSQ performed well in higher-precision (e.g., 4-bit quantization), its effectiveness degrades significantly in the extreme low-bit regimes. To address this, SEQ employs a more symmetric quantization grid, leading to improved performance. Inference scaling laws for models w.r.t quantization have been extensively studied in \cite{dettmers2023case}, \cite{ vsubic2026balancing}, and \cite{paretoq}. 

BitNet was proposed in \cite{bitnet158hf}, which introduced the idea of gradually transitioning model weights towards the target low-bit representation through a predefined schedule (e.g., linear schedule). This is accomplished by linearly combining the latent weights with appropriately scaled version (based on a schedule) of difference of FP latent weight and the desired quantized weight. However, in our proposed RelaxQAT, the bitwidth itself changes based on a schedule. Other follow up works in Bitnet w.r.t activation quantization, 1-bit optimzation and efficient inference are detailed in \cite{wang2024bitnet}, \cite{wang2025bitnetLLM}, \cite{wang2025bitnetedge}. 

\cite{winq} investigates the root cause of  slow convergence observed in quantization-aware training (QAT) for extreme low bit quantization. It is shown that low-bit training tends to become trapped in flat regions around saddle points of the loss landscape, resulting in slower optimization. To mitigate this issue, an optimization strategy is proposed that periodically interpolates the quantized weights with the latent weights to escape the saddle point. Further, \cite{xu2025butterflyquant}, \cite{xu2026bit},\cite{wei2022qdrop}, \cite{lv2026makes}, \cite{shao2024omniquant}, \cite{lee2024owq}, \cite{wang2025outlier}, \cite{wang2025ofq} analyse several low-bit training methods and how outlier-aware quantization schedules improve convergence and accuracy at extreme bit widths.

A direction  closely related to our proposal is the progressive quantization framework introduced in Unifying Block-wise PTQ and Distillation-based QAT for Progressive Quantization (UPQ) \cite{upq}. UPQ proposes a two-stage quantization pipeline in which a full-precision model is  converted to an intermediate  representation (e.g., INT4) using block-wise post-training quantization (PTQ), followed by INT2 QAT initialized from the intermediate checkpoint. In contrast, the proposed RelaxQAT generalizes the notion of progressive quantization by replacing the fixed transition (FP to INT4 to INT2) with a continuously evolving bit schedule. We hypothesize that this continuous transition enables a smoother optimization trajectory as shown next.

\section{Proposed RelaxQAT}\label{sec::approach}
RelaxQAT consists of two parts: (\textit{i})~set bitwidth using a smooth schedule during training, and (\textit{ii})~modify existing quantization equations such as Stretched Elastic Quantizer (SEQ)~\cite{paretoq} to make sure the extreme values of the weight distribution become floating points so they can smoothly vary as the bitwidth changes. We describe these components below.

\subsection{Bitwidth Schedule}\label{sec:bws}
The bitwidth ($b_t$) at training step $t$ is given by:
\begin{equation}
    b_t = b_s - (b_s - b_e) \text{sin}\left(\frac{\pi}{2} \omega \right), \ \ \ \ \omega \in [0,1]
    \label{eq:bws}
\end{equation}
Here, $b_s$ and $b_e$ are the starting and ending bitwidths, respectively, and $\omega\in[0,1]$ dynamically changes during training.  Eq.~\eqref{eq:bws} states that when $\omega=0$, we get the starting bitwidth ($b_s$), and when $\omega=1$, we get the desired final bitwidth ($b_e$). Typically, the $\omega$ starts at 0 and ends at 1 and then follows a cyclical pattern. Before the end of the training, we keep $\omega=1$ to ensure that the model finetunes in the desired $b_e$ bitwidth for a prolonged time.

\begin{figure}[t]
  \centering
   \includegraphics[width=1\linewidth]{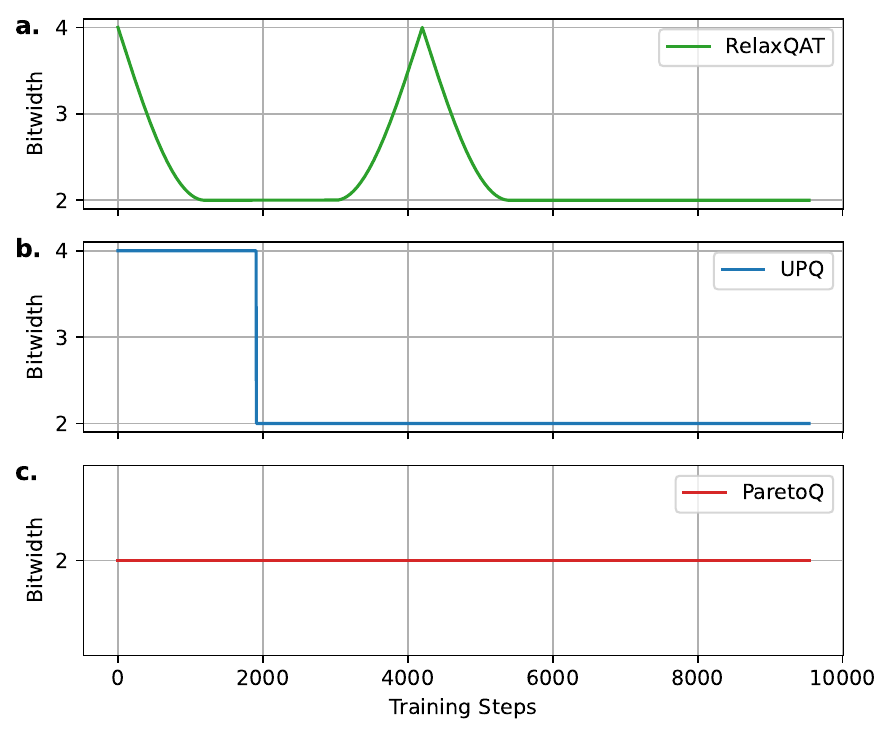}\vspace{-6mm}
   \caption{(a)~An example of our proposed RelaxQAT bitwidth schedule which gradually and smoothly reduces the bitwidth. The gradient of the schedule 0 whenever it is approaches 2-bits; this helps to ease into the 2-bit precision instead of an abrupt change. (b)~Prior progressive quantization methods reduce the bitwidth suddenly during training like UPQ, or (c)~keep it the same throughout.}\vspace{-1mm}
   \label{fig:bws}
\end{figure}

An example of our proposed bitwidth schedule is shown in Fig.~\ref{fig:bws}(a). We typically start with a \textit{half cycle} going from bitwidth $b_s$ to $b_e$ over \textit{maximum schedule steps} ($m_s$ training steps). For instance, in Fig.~\ref{fig:bws}(a), $b_s=4$, $b_e=2$, $m_s=1200$). This corresponds to $\omega$ changing from 0 to 1. After this half cycle, we can have $n$ \textit{full cycles} ($n\in \{1,2,3,...,\ldots,N\}$) where the $\omega$ cycles from 1 to 0 and then back to 1. This results in bitwidth changing from $b_e$ to $b_s$ and then back to $b_e$. As shown in Fig.~\ref{fig:bws}(a), at step 3000, the quantization bitwidth starts to relax from 2 bits to 4 bits (by step 4200, since $m_s=1200$ and $n=1$) and then goes back to 2 bits by step 5400. The intuition behind adding one or more full cycles after the initial half cycle is to help the optimization escape any flat regions that it might be stuck in. We empirically validate that the additional full cycles indeed help improve the RelaxQAT performance.

It is important to note that Eq.~\eqref{eq:bws} leads to a slowly changing bit precision which can be a floating point value. Moreover, the gradient of the proposed schedule is 0 whenever it approaches 2-bits. This helps us to ease into 2-bit precision instead of a sudden change in bitwidth. In contrast, existing methods either (\textit{i})~perform progressive quantization to train the model in stages and abruptly change the bitwidth (e.g., UPQ~\cite{upq}, see Fig.~\ref{fig:bws}(b) blue line around step 2000), or (\textit{ii})~keep it the same throughout the QAT after directly reducing it from the floating point precision (e.g., ParetoQ~\cite{paretoq}, see Fig~\ref{fig:bws}(c) red line). 

Our hypothesis is that the relaxation in bitwidth can help the optimization escape the flat regions in the quantization loss surface. Further, a sudden change in bitwidth in methods such as UPQ can cause the loss to increase suddenly, and may need some training steps to recover it. Since RelaxQAT has a slowly changing floating point bitwidth, this reduces \textit{sudden} spikes in the loss function. Next, we explain the changes to the quantization equations needed to realize the proposed floating point bitwidth.

\subsection{RelaxSEQ}\label{sec::relaxseq}
QAT first replaces each model weight with its  corresponding quantized-dequantized (QDQ) weight which is computed as a function of a latent floating point weight $\theta$ and its channelwise scale $\alpha$. Specifically, given a $\theta\in\mathbb{R}^{n\times m}$, the latent floating point weight with $n$ output channels and $m$ input channels and a channelwise scale $\alpha\in\mathbb{R}^n$, various quantization schemes compute the QDQ weights in different ways~\cite{lsq, paretoq}. Then, the model is trained with the latent weight $\theta$ and the scale $\alpha$ as the trainable parameters, where the QDQ weight is computed in every forward pass and passed on to the model layers. The associated backward pass is also modified using Straight-Through Estimators (STE), etc.~\cite{lsq}. 

Recently, a popular 2-bit and sub-2-bit (e.g., ternary, 1.58-bits) scheme called Stretched Elastic Quantizer (SEQ) was proposed in ParetoQ~\cite{paretoq}. The main advantage of SEQ compared to LSQ~\cite{lsq} for 2-bit QAT is that it results in a symmetric $\{-3, -1, 1, 3\}$ integer grid. In contrast, LSQ results in an aysmmetric $\{-2, -1, 0, 1\}$ integer grid which does not perform as well for 2-bit QAT. The QDQ weight $w_p$ using the existing SEQ~\cite{paretoq} is computed as:
\begin{equation}
    w_p = \frac{\alpha}{k} \left\{ \text{round}\left[\text{clip}\left(\frac{\theta}{\alpha}, -1, 1 \right) \times k - 0.5 \right] + 0.5 \right\}.
    \label{eq:orig_seq}
\end{equation}
Here, $k= 2^{b-1}$, $b$ is the number of bits. In principle, we can use Eq.~\ref{eq:bws} to assign a dynamically changing, floating point bitwidth to the existing SEQ Eq.~\eqref{eq:orig_seq} because $k$ is a function of the number of bits. That is, the range of the QDQ weight distribution simply expands or shrinks as $b$ changes. Note that, $w_p$ first performs a clipping and then rounds the resulting expression. Thus, excluding the $\alpha/k$ scaling and the $+0.5$ shift in Eq.~\eqref{eq:orig_seq}, this \textit{Clip-then-Round} operation obviously results in \textit{integer} values for all the quantization bins, including the \textit{extremes}. We next analyze how these extreme quantization bins vary when the bitwidth changes.

\begin{figure}[t]
  \centering
   \includegraphics[width=0.8\linewidth]{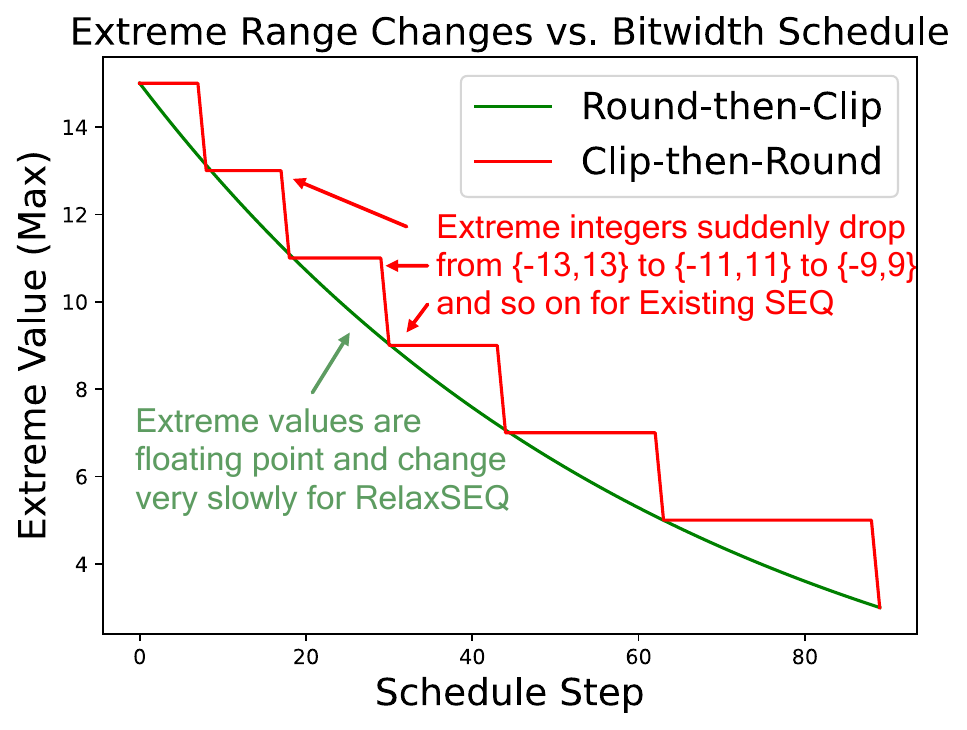}\vspace{-5mm}
   \caption{ParetoQ SEQ with a bitwidth schedule loses extreme integers as $k=2^{b_t-1}$ decreases with a shrinking bitwidth. RelaxSEQ shows a smooth change in extreme values which results in a smoothly changing loss.}
   \label{fig:extreme}
\end{figure}

Suppose that we are given a latent weight $\theta\sim \mathcal{N}(0, 0.02)$ and its corresponding channelwise scale $\alpha = \text{max}\{\theta\}$. For the purpose of this simulation, the latent weight and the scale are fixed, and only the bitwidth is reduced from 4 bits to 2 bits in Eq.~\eqref{eq:orig_seq}. In Fig.~\ref{fig:extreme}, we plot how the extreme values of the distribution change as the schedule continues to reduce the bitwidth. Clearly, the existing SEQ (Clip-then-Round, {\color{red} red}) shows a sudden, staircase drop in extreme integers. 

Why does this sudden change in integer ranges pose a problem? During QAT, the training continues between the staircase drops assuming that those extreme integers are available. However, when $b_t$ (and thus $k=2^{b_t-1}$) is reduced due to a schedule, those extreme integers suddenly disappear. In other words, during training, if the bitwidth is dynamically reduced, the model weights get finetuned assuming that we have these extreme integers. However, these extreme integers are abruptly lost as the training goes on. This loss of extreme integers can result in unstable training behavior and \textit{sudden} loss spikes (similar to UPQ~\cite{upq}). Indeed, additional training steps are needed to recover from each loss spike. Consequently, we need to modify the existing SEQ to gradually change the extreme values in the QDQ distribution. To this end, we propose \textit{RelaxSEQ} as: 
\begin{equation}
    w_r = \frac{\alpha}{k} \left\{ \text{clip}\left(\text{round}\left[ \frac{\theta}{\alpha} \times k - 0.5 \right], -k, k-1 \right) + 0.5 \right\}.
    \label{eq:relax_seq}
\end{equation}

Essentially, RelaxSEQ flips the order of operations from Clip-then-Round to \textit{Round-then-Clip} and makes the clipping ranges a function of $k$. Therefore, when the bitwidth is floating point, the extreme values (resulting from the clipping operation) are also floating point. Meanwhile, similar to the existing SEQ, the distribution can still shrink and expand with the changing bitwidth. As a result, when the precision changes, there is a smooth change in the extreme values of the distribution, which leads to a smoothly changing loss with no sudden spikes. As $\omega$ becomes 1 towards the end of the training, the model converges to the desired low-bitwidth. Fig.~\ref{fig:extreme} ({\color{ForestGreen} green}) demonstrates that as the bit schedule progresses, the extreme values smoothly change instead of a sudden staircase drop as in the existing SEQ. More details on existing SEQ vs. RelaxSEQ including some early training and benchmarking results are shown in Appendix~\ref{app:extremeval}, Figs.~\ref{fig:lossExSEQNewSEQ},~\ref{fig:ctrrtc}, and Table~\ref{tab:mmluNewSeqOldSeq}.

\subsection{RelaxQAT}
We now combine the bit schedule and the RelaxSEQ to get the proposed RelaxQAT method. We implement the bit schedule as a \texttt{transformers.TrainerCallback}. Algorithm~\ref{alg:callback-init} shows a psuedo-code to initialize the bitwidth
\begin{algorithm}[t]
\caption{RelaxQATCallback: At initialization}
\label{alg:callback-init}
\begin{algorithmic}[1]
\Function{InitializeRelaxQATCallback}{model, start\_bits ($b_s$), end\_bits ($b_e$), cycle\_start\_step ($c_{ss}$), cycle\_end\_step ($c_{es}$), cycle\_length ($c_l$), max\_schedule\_steps ($m_s$)}
    \State $\operatorname{Assert}$ $c_{ss} >$ $m_s$
    \State $\operatorname{Assert}$ $c_l> 2 \cdot m_s$
    \State $\operatorname{Assert}$ $b_s > b_e$
    \State $\operatorname{Assert}$ $(c_{es} - c_{ss}) \% c_l = 0$
    \State $\omega \gets \operatorname{linspace}(0, 1, \text{num}=m_s)$
    \State $B \gets (b_s - (b_s - b_e)$
        $\sin((\pi/2) \omega))$
    \State $n \gets ((c_{es}-c_{ss})/c_l)+1$\ \ \ \ \ \ \ \ \ \ \ \ \  \# number of cycles
    \State $\mathcal{C} \gets \operatorname{linspace}(c_{ss}, c_{es}, \text{num}=n)$
    
    \State $modules \gets$ all \texttt{SEQ} modules in model
    
    \ForAll{$m \in modules$}
        \State $m.enable\_relax\_qat \gets \textbf{true}$
        \State $m.i \gets 0$\ \ \ \ \ \ \ \ \ \ \ \ \ \ \ \ \ \ \  \# current bit schedule index
        \State $m.B \gets B$\ \ \ \ \ \ \ \ \ \ \ \ \ \ \ \ \ \ \ \ \# bit schedule list: $[b_s, b_e]$
        \State $m.cci \gets 0$\ \ \ \ \ \ \ \ \ \ \ \ \ \ \ \ \ \ \ \ \ \ \ \ \ \ \ \ \ \ \ \ \ \ \# current\_cycle\_idx
        \State $m.pci \gets -1$\ \ \ \ \ \ \ \ \ \ \ \ \ \ \ \ \ \ \ \ \ \ \ \ \ \ \ \ \# previous\_cycle\_idx
    \EndFor
    \State \Return initialized callback state
\EndFunction
\end{algorithmic}
\end{algorithm}
schedule. It requires start bits ($b_s$), end bits ($b_e$), cycle start step ($c_{ss}$), cycle end step ($c_{es}$), cycle length ($c_l$), and maximum schedule steps ($m_s$) as arguments. Then,  an array $\omega$ is initialized to evenly spaced $m_s$ numbers between 0 and 1 (all inclusive). Eq.~\eqref{eq:bws} is then used to obtain a list of floating point bitwidths $B$. The number of \textit{full cycles} $n$ and cycle intervals $\mathcal{C}$ are obtained using $c_{ss}$, $c_{es}$, and $c_l$. Finally, for each SEQ module, various states are initialized (e.g., cycle indices, bit schedule list, bit schedule index).

\begin{algorithm}[t]
\caption{RelaxQATCallback: At each training step}
\label{alg:callback-step}
\begin{algorithmic}[1]
\Function{RelaxQATOnStepBegin}{$modules$, global\_step ($t$), cycle\_intervals ($\mathcal{C}$), max\_schedule\_steps ($m_s$)}
    \ForAll{$m \in modules$}
        \If{$t < \mathcal{C}[0]$} \ \ \ \# first half cycle from $b_s$ to $b_e$
            \State $m.i \gets \min(t, m_s - 1)$
        \ElsIf{$t < \mathcal{C}[-1]$} \ \ \ \ \ \ \ \ \ \ \ \ \ \ \ \ \ \ \ \# $n$ full cycles
            \If{$t = \mathcal{C}[m.cci]$}
                \State \# Update current and previous cycle idx
                \State $m.i \gets m_s - 1$
                \State $m.cci \gets m.cci + 1$
                \State $m.pci \gets m.pci + 1$
            \ElsIf{$t \leq \mathcal{C}[m.pci]
                + m_s - 1$}
                \State \# first: relax bits from $b_e$ to $b_s$
                \State $m.i \gets \mathcal{C}[m.pci] + m_s - 1 - t$
            \Else
                \State \# second: reduce bits from $b_s$ to $b_e$
                \State $m.i \gets \min(
                    t - (\mathcal{C}[m.pci]
                    + m_s - 1),
                    m_s - 1)$
            \EndIf
        \Else \ \ \ \ \ \ \ \ \ \ \ \ \ \ \ \ \ \ \# all $n$ cycles ended: set bits to $b_e$
            \State $m.i \gets m_s - 1$
        \EndIf
        \State $m.b_t \gets m.B[m.i]$ \ \ \# set dynamic bitwidth $b_t$
    \EndFor
    \State \Return dynamic bitwidth $b_t$ for all \texttt{SEQ} modules
\EndFunction
\end{algorithmic}
\end{algorithm}

After initialization, we compute the dynamic bitwidth at each training step as shown in Algorithm~\ref{alg:callback-step}. As mentioned in Section~\ref{sec:bws}, we start with a half cycle going from $b_s$ to $b_e$ bits. Then we conduct $n$ full cycles going from $b_e$ to $b_s$ and then back to $b_e$ bitwidth. Once all $n$ cycles have ended, we set the bitwidth to $b_e$ for the final finetuning in the desired bit precision. Hence, at each global step $t$ during training, the dynamic bitwidth $b_t$ is set using Algorithm~\ref{alg:callback-step}. Finally, we assign the dynamic bitwidth $b_t$ to the RelaxSEQ function. The forward function shown in Algorithm~\ref{alg:relaxseq} computes the QDQ weight $w_r$ and sends it to the \texttt{nn.Linear} layers in the transformer modules. The backward function for RelaxSEQ is similar to the backward function of existing SEQ~\cite{paretoq}, except the clipping ranges are changed based on Eq.~\eqref{eq:relax_seq}. This completes the description of our proposed RelaxQAT technique.

\begin{algorithm}[t]
\caption{RelaxSEQ for 2-bit QAT}
\label{alg:relaxseq}
\begin{algorithmic}[1]
\Function{RelaxSEQ}{$\theta, \alpha, b$}
    \State $n_{levels} \gets 2^{b - 1}$
    \State shift$ \gets 0.5$
    \State $\delta \gets \alpha / n_{levels}$
    \State $q_{bins} \gets \operatorname{round}(\theta / \delta - $shift$)$
    \State $q_{bins} \gets \operatorname{clip}(q_{bins}, -n_{levels}, n_{levels} - 1)$
    \State $q \gets q_{bins} + $shift
    \State $w_r \gets q \cdot \delta$
    \State \Return $w_r$
\EndFunction
\end{algorithmic}
\end{algorithm}

\subsection{Systems Considerations}
Any new QAT paradigm must not come at the expense of training speed, particularly because low-bit (2-bit or 
sub-2-bit) QAT requires billions of tokens in training in order to recover sufficient accuracy. To this end, we implemented triton kernels~\cite{tillet2019triton} for both Re-
\begin{wraptable}{r}{4.5cm}
    \scalebox{1.0}{
    \addtolength{\tabcolsep}{-2pt}
    \fontsize{8.0pt}{5.75pt}\selectfont
    \begin{tabular}{lc}
    \toprule
    \textbf{Method} & \textbf{GPU Hours} \\
    \midrule
    ParetoQ & 370 \\
    \midrule
    \textbf{RelaxQAT (ours)} & \textbf{361 {\color{ForestGreen}(-2.4\%)}} \\
    
    \bottomrule
    \end{tabular}\vspace{-5mm}
    }
    \caption{Training Speeds for Qwen3-0.6B model on 10B tokens.} 
    \label{tab:trainspeed}
\end{wraptable} 
laxSEQ and the existing ParetoQ SEQ functions. 
We train Qwen3-0.6B model~\cite{qwen3} on 10B tokens on eight NVIDIA H100 GPUs. Table~\ref{tab:trainspeed} shows the end-to-end training time for ParetoQ and RelaxQAT in GPU hours. As evident, both ParetoQ and RelaxQAT achieve very similar GPU hours with RelaxQAT being slightly faster (but within margin of error). Note that, Algorithms~\ref{alg:callback-init} and~\ref{alg:callback-step} precompute the bit schedule at initialization. Hence, the bitwdith is simply assigned to RelaxSEQ without any computation of Eq.~\eqref{eq:bws} during the forward pass. This further removes any additional overhead resulting from the bit schedule in RelaxQAT.

\section{Theoretical Insights}\label{sec::theory}
Let us now derive some simple theoretical insights for RelaxQAT. Although RelaxQAT is proposed for SEQ, the same principles apply to other quantizers such as Learned Step-size Quantizer (LSQ)~\cite{lsq}. The LSQ QDQ weight is computed as:
\begin{equation}
    w_l = \frac{\alpha}{k} \left\{ \text{clip}\left(\text{round}\left[ \frac{\theta}{\alpha} \times k\right], -k, k-1 \right) \right\}.
    \label{eq:lsq}
\end{equation}

Once again, $k=2^{b-1}$. It is easy to see why the same RelaxQAT ideas apply to LSQ. The main difference between Eq.~\eqref{eq:relax_seq} and~\eqref{eq:lsq} is the shift factor of 0.5 which changes the integer grid from $\{-3, -1, 1, 3\}$ for RelaxQAT to $\{-2, -1, 0, 1\}$ for LSQ for a 2-bit QAT. Since the LSQ grid is simpler to analyze, without loss of generality, we will use it in this section to derive our theoretical insights.

We will show two key results: (\textit{i})~How is bitwidth related to the flatness of the quantization loss surface and how we may need integer changes in QDQ weights to make progress when we are stuck on a flat surface? (\textit{ii})~Why does the optimization stall when the QAT does not result in a substantial change in integer values? We will also demonstrate how RelaxQAT can help with the above problems.
\begin{prop}
The higher the bit precision, the smaller the width of the quantization hyperplanes. Specifically, given two bitwidths $b_1$ and $b_2$, with $b_1 < b_2$, the width of the quantization hyperplanes can be reduced by a factor of $2^{b_2-b_1}$ if we relax the bitwidth from $b_1$ to $b_2$. 

Therefore, relaxing the bitwidth during QAT can help the optimization escape the flat regions in the loss surface.
\label{prop1}
\end{prop}
\begin{proof}
See Appendix~\ref{app:prop1proof} for the proof.
\end{proof}

\begin{figure}[t]
  \centering
   \includegraphics[width=1\linewidth]{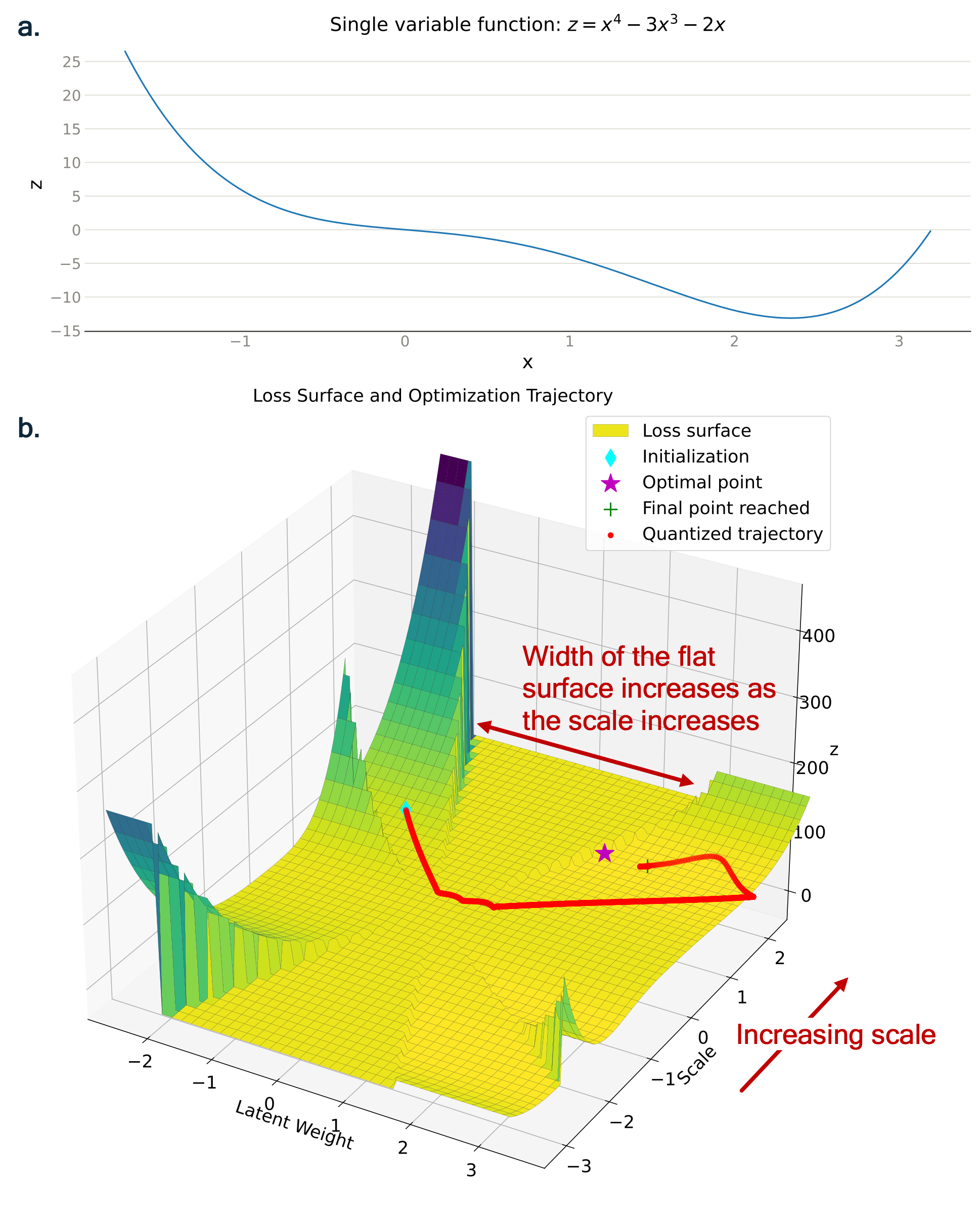}\vspace{-5mm}
   \caption{(a)~A simple 1-d function to minimize. (b)~The 1-d function above results in a highly complex 2-d, 2-bit quantization loss surface (generated using LSQ) where the width of the hyperplanes (defined as the flat regions between integer changes) increases with larger scales. Red line shows the QAT trajectory.}
   \label{fig:lt}
\end{figure}

We next simulate a real quantization loss surface for a toy problem to illustrate how flatness increases with larger scale values. Fig.~\ref{fig:lt}(a) plots a simple 1-d function: $x^4-3x^3-2x$. The problem is to minimize this function. Now, if we choose $x$ as the optimization variable, a 1-d function in the floating point domain will result in a 2-d optimization problem in the quantized realm. This is because the scalar $x$ splits into a scalar latent weight $\theta$ and a scalar $\delta$ scale. Thus, the minimization problem is ${w_l}^4-3{w_l}^3-2{w_l}$, where $w_l$ comes from the LSQ Eq.~\eqref{eq:lsq} and $\{\theta, \delta\}$ are the two optimization variables. Fig.~\ref{fig:lt}(b) shows the highly complex 2-d, 2-bit quantization loss surface arising from a simple 1-d minimization problem. As evident, we see clear quantization hyperplanes marking the integer value changes in the QDQ weight. More importantly, we see that as the scale becomes larger, the width of the hyperplanes increases, just as predicted in Proposition~\ref{prop1}. Hence, relaxing the bitwidth (which reduces the scale $\delta$) can help a stuck optimization by reducing the width of such hyperplanes.

Fig.~\ref{fig:lt}(b) also shows a red line tracing the optimization path on the quantized loss surface. It moves across multiple hyperplanes, which show changes in the integer value in the QDQ weight, to reach near the minimum point. An interesting problem then would be to understand how important these integer changes are for a QAT. Moreover, can we mathematically capture the QAT optimization trajectory (e.g., red line in Fig.~\ref{fig:lt}(b))? This leads us to our next theoretical result.

\begin{prop}
The direction of the QAT trajectory is primarily decided by the scale gradients only and not by the latent weight gradients if there are no integer changes in the QDQ weight tensor. This can slow down the training convergence since there are a lot more latent weights than (channelwise) scale parameters in the model.

Since RelaxQAT relaxes the bitwidth which reduces the width of quantization hyperplanes defined by the integer-change boundaries, it makes it easier to change integers. This directly contributes to the directions the QAT trajectory can take by better utilizing the latent weight gradients, and thus results in faster convergence. 
\label{prop2}
\end{prop}
\begin{proof}
Let us first rewrite Eq.~\eqref{eq:lsq} in the following form:
\begin{equation}
    w^{(t)} = \delta^{(t)} \left\lfloor\frac{\theta^{(t)}}{\delta^{(t)}}\right\rceil = \delta^{(t)} z^{(t)},
    \label{eq:anotherlsq}
\end{equation}
where, $w^{(t)}$ is the QDQ weight at training step $t$, $\delta^{(t)}=\frac{\alpha^{(t)}}{2^{b-1}}$ is the scale at step $t$, and $b$ is an \textit{integer} bitwidth, e.g., 2-bits. The $\lfloor\rceil$ operator should be understood as the Round-then-Clip from Eq.~\eqref{eq:lsq}. Since $b$ is an integer here, $z^{(t)}$ is the integer value of the QDQ weight at step $t$.

Now, let us mathematically describe the QAT trajectory (red line in Fig.~\ref{fig:lt}(b)). The QDQ weight at step $t+1$ is given by:
\begin{equation}
    \begin{aligned}
    w^{(t+1)} &=  \delta^{(t+1)} \left\lfloor\frac{\theta^{(t+1)}}{\delta^{(t+1)}}\right\rceil\\
    &= (\delta^{(t)} - \eta d_\delta^{(t)})\left\lfloor\frac{(\theta^{(t)} - \eta d_\theta^{(t)})}{(\delta^{(t)} - \eta d_\delta^{(t)})}\right\rceil\\
    &= (\delta^{(t)} - \eta d_\delta^{(t)})z^{(t+1)}\\
    &= \delta^{(t)}z^{(t+1)} - \eta z^{(t+1)}d_\delta^{(t)},
    \end{aligned}
    \label{eq:qdq}
\end{equation}
where, $\eta$ is the learning rate, $d_\delta^{(t)}$ ($d_\theta^{(t)}$) is the optimizer (e.g., Adam or SGD) update as a function of the gradient of $\delta$ ($\theta$), and $z^{(t+1)}$ is the integer value at step $t+1$. Note that, the gradient information of $\theta$ is contained only within $z^{(t+1)}$ and does not appear anywhere else in Eq.~\eqref{eq:qdq}. When there is no integer change at a given training step, we have:
\begin{equation}
    \begin{aligned}
    w^{(t+1)} &=  \delta^{(t)}z^{(t+1)} - \eta z^{(t+1)}d_\delta^{(t)}\\
    &= \delta^{(t)}z^{(t)} - \eta z^{(t)}d_\delta^{(t)}\ \ \ \ \ (\text{since }z^{(t+1)} = z^{(t)})\\
    &= w^{(t)} - \eta z^{(t)}d_\delta^{(t)}\ \ \ \ \ \ \ \ \ \ (\text{Eq.~\eqref{eq:anotherlsq}})\\
    &= w^{(t)} - \eta d_w^{(t)},
    \end{aligned}
    \label{eq:nointchange}
\end{equation}
where, $d_w^{(t)} = z^{(t)}d_\delta^{(t)}$. Hence, when the integer values $z$ do not change, there are no gradient directions ($d_\theta^{(t)}$) coming from the latent weights, and the optimization direction is primarily driven by the scales. Further, because the number of latent weights far exceed the (channelwise) scales, a lack of integer changes during QAT can slow down the training convergence. This highlights how gradual changes to the loss surface induced by RelaxQAT can lead to more integer changes, and thus meaningfully contribute to the QAT convergence by better utilizing the latent weight gradients.
\end{proof}

Finally, similar to Fig.~\ref{fig:lt}(b) which was produced using LSQ, we repeat the analysis for the ParetoQ SEQ and RelaxSEQ (Eqs.~\eqref{eq:orig_seq} and~\eqref{eq:relax_seq}, respectively). The loss visualization is shown in Fig.~\ref{fig:losssim} in Appendix~\ref{app:seqsim}. 

\section{Experiments}\label{sec::exp}

\subsection{Experimental Setup}

We perform experiments on Qwen3 (0.6B/4B), Qwen3.5 (2B/4B) and Gemma3-1B models in the language domain and Qwen3.5 (2B/4B) models in the vision-language domain. We compare our method against five state-of-the-art QAT baselines, including ParetoQ~\cite{paretoq}, LLM-QAT~\cite{llmqat}, UPQ~\cite{upq}, BitNet~\cite{bitnet158hf}, and WinQ~\cite{winq}. Additionally, we evaluate AdaScale~\cite{adascaleptq}, a PTQ method for comparison. For BitNet, since~\cite{bitnet158hf} is only for 1.58 bits and has no learnable scale, we adopt ParetoQ's SEQ function to get the 2-bit QDQ weight and use it inside the BitNet algorithm. 

The training data consists of a mixture of DCLM-Edu ~\cite{allal2025smollm2smolgoesbig}, tulu-3-sft-olmo-2-mixture ~\cite{tulu3sftolmo2mixture} and xlam-function-calling-60k ~\cite{liu2024apigen}. For vision-language model training, we additionally include Llama-Nemotron-VLM-Dataset-v1 ~\cite{llamanemotronvlmdataset}. We train all models on 10B tokens. We use the same initial learning rate of 0.0001 for all  models and baselines with a cosine learning rate schedule that decays to nearly zero. The smaller models are trained on a single node of eight H100 GPUs whereas the bigger models are trained on two A100 nodes. Note that, we only do weight only quantization in all experiments.
\begin{table*}[t] 
    \centering 
    \scalebox{0.92}{ 
    \setlength{\tabcolsep}{6pt} 
    \renewcommand{\arraystretch}{1.2} 
    \fontsize{8.0pt}{8.5pt}\selectfont 
    \begin{tabular}{l|c||ccccc||c|c|c||c} \hline \textbf{Method} & \textbf{Type} & \textbf{ArcE($\uparrow$)} & \textbf{ArcC($\uparrow$)} & \textbf{HS($\uparrow$)} & \textbf{WG($\uparrow$)} & \textbf{PIQA($\uparrow$)} & \textbf{CSR($\uparrow$)} & \textbf{MMLU($\uparrow$)} & \textbf{IFEval($\uparrow$)} & \textbf{Global Avg.($\uparrow$)} \\ 
    \hline 
    \noalign{\vskip 0.5pt} 
    \hline 
    FP & - & 81.22 & 58.44 & 69.08 & 67.48 & 75.73 & 70.39 & 72.63 & 86.55 & 76.52 \\ 
    \hline AdaScale~\cite{adascaleptq} & PTQ & 25.54 & 26.36 & 26.94 & 50.19 & 50.76 & 35.95 & 26.89 & 20.27 & 27.70 \\ 
    \hline ParetoQ~\cite{paretoq} & \multirow{6}{*}{QAT} & 74.92 & 48.38 & 62.08 & 63.77 & 73.78 & 64.58 & 62.30 & 80.22 & 69.03{\color{ForestGreen}(+0.00\%)} \\ 
    UPQ~\cite{upq} & & 75.97 & \underline{49.66} & 61.79 & 62.67 & 73.23 & 64.66 & 61.21 & 79.18 & 68.35 {\color{red}(-0.68\%)}\\
    LLM-QAT~\cite{llmqat} & & \textbf{76.94} & 48.29 & 61.29 & 63.93 & 73.45 & 64.78 & 59.66 & 79.73 & 68.05 {\color{red}(-0.98\%)}\\ 
    BitNet~\cite{bitnet158hf} & & 69.86 & 43.51 & 60.67 & 61.95 & 71.27 & 61.45 & 60.15 & 79.21 & 66.93 {\color{red}(-2.10\%)}\\
    WinQ~\cite{winq} & & 75.42 & 48.55 & \underline{62.23} & \textbf{65.51} & \underline{73.99} & \underline{65.14} & \underline{62.80} & \underline{80.26} & \underline{69.40} {\color{ForestGreen}(+0.37\%)} \\ 
    \textbf{RelaxQAT (ours)} & & \underline{76.47} & \textbf{50.26} & \textbf{63.05} & \underline{64.48} & \textbf{74.43} & \textbf{65.73} & \textbf{62.84} & \textbf{81.54} & \textbf{70.03} {\color{ForestGreen}\textbf{(+1.00\%)}}\\ \hline 
    \end{tabular} } 
    \vspace{-3mm} 
    \caption{Evaluation of Qwen3-4B on Commonsense Reasoning, MMLU, and IF-Eval benchmarks. WG and HS denote WinoGrande and HellaSwag, respectively. ($\uparrow$): higher is better. {\color{ForestGreen} Green} and {\color{red} red} denote improvement and degradation relative to ParetoQ, respectively. \textbf{Bold} is best and \underline{underline} is second-best.} \label{tab:q34bllm} 
\end{table*}

We select $n=1$ full cycle (after an initial half cycle) for all experiments. For maximum schedule steps $m_s$, we perform a grid search between $\{400, 800, 1200\}$. We also conduct a detailed ablation study to show that RelaxQAT is robust across hyperparameter choices for $n$ and $m_s$. Also, we set $b_s=4$ and $b_e=2$ bits, signifying a 4-bit to 2-bit schedule.  

In the language domain, we evaluate RelaxQAT on five zero-shot common sense reasoning tasks. These include Arc-easy, Arc-challenge ~\cite{clark2018think}, HellaSwag ~\cite{zellers2019hellaswag}, Winogrande ~\cite{sakaguchi2021winogrande}, and PIQA ~\cite{bisk2020piqa}. We also report five-shot MMLU ~\cite{hendrycks2020measuring} (loglikelihood task) and IFEval ~\cite{zhou2023instruction} (generative task) benchmarks, each focusing on evaluating the knowledge and instruction following capabilities of the model. In vision domain, we evaluate on MMMU ~\cite{yue2024mmmu}, DocVQA ~\cite{mathew2021docvqa}, and ScienceQA ~\cite{lu2022learn} benchmarks.

\subsection{Language Modeling Results}
\begin{figure}[t]
  \centering
   \includegraphics[width=0.85\linewidth]{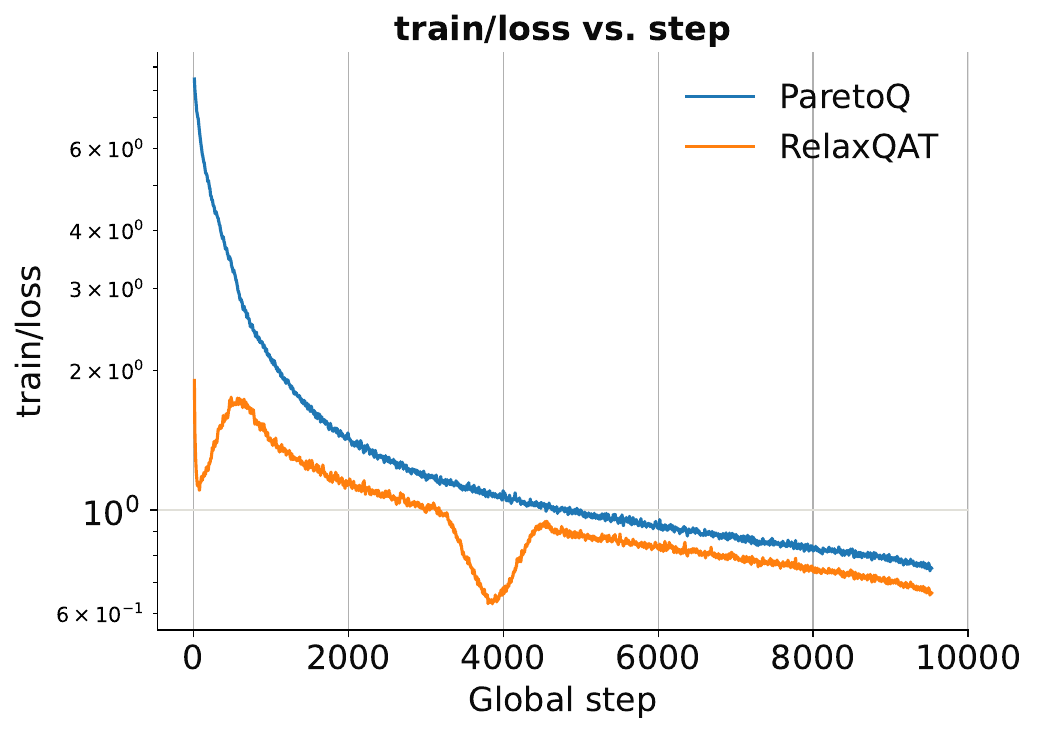}\vspace{-5mm}
   \caption{RelaxQAT converges significantly faster than ParetoQ.}
   \label{fig:tl}
\end{figure}

We first look at the loss convergence of RelaxQAT. Fig.~\ref{fig:tl} demonstrates the training convergence of RelaxQAT vs. ParetoQ for the Qwen3-0.6B model class. As evident, the initial half cycle from 4-bits to 2-bits significantly boosts the accuracy. However, by step 3000 or so, ParetoQ starts to catch up. The final ($n=1$) full cycle further accelerates the RelaxQAT towards the end. Hence, RelaxQAT converges faster and achieves a significantly better training loss.
\begin{table*}[t] 
    \centering 
    \scalebox{0.92}{ \setlength{\tabcolsep}{6pt} 
    \renewcommand{\arraystretch}{1.2} 
    \fontsize{8.0pt}{8.5pt}\selectfont 
    \begin{tabular}{l|c||ccccc||c|c|c||c} 
    \hline 
    \textbf{Method} & \textbf{Type} & \textbf{ArcE($\uparrow$)} & \textbf{ArcC($\uparrow$)} & \textbf{HS($\uparrow$)} & \textbf{WG($\uparrow$)} & \textbf{PIQA($\uparrow$)} & \textbf{CSR($\uparrow$)} & \textbf{MMLU($\uparrow$)} & \textbf{IFEval($\uparrow$)} & \textbf{Global Avg.($\uparrow$)} \\ 
    \hline 
    \noalign{\vskip 0.5pt} 
    \hline 
    FP & - & 56.01 & 34.55 & 47.29 & 55.48 & 67.73 & 52.21 & 47.25 & 64.78 & 54.74 \\ 
    \hline 
    AdaScale~\cite{adascaleptq} & PTQ & 40.45 & 25.51 & 32.78 & 52.49 & 59.41 & 42.12 & 25.96 & 21 & 29.69 \\ 
    \hline 
    ParetoQ~\cite{paretoq} & \multirow{6}{*}{QAT} & 50.79 & 27.64 & 40.19 & 52.40 & 64.03 & 47.01 & 37.18 & 33.85 & 39.34 {\color{ForestGreen}(+0.00\%)} \\ 
    UPQ~\cite{upq} & & 50.42 & 27.56 & 39.58 & 52.09 & 63.82 & 46.69 & 36.54 & 43.89 & 42.37 {\color{ForestGreen}(+3.03\%)} \\
    LLM-QAT~\cite{llmqat} & & 49.91 & 28.41 & 39.76 & 52.72 & \underline{64.09} & 46.97 & 37.35 & 44.61 & 42.97 {\color{ForestGreen}(+3.63\%)} \\ 
    BitNet~\cite{bitnet158hf} & & \underline{52.44} & \underline{28.49} & \underline{40.34} & \textbf{55.64} & \textbf{64.20} & \underline{48.22} & \underline{37.62} & 49.29 & \underline{45.04} {\color{ForestGreen}(+5.70\%)} \\ 
    WinQ~\cite{winq} & & 50.16 & 27.30 & 40.12 & \underline{55.01} & 63.76 & 47.27 & 37.02 & \underline{49.56} & 44.61 {\color{ForestGreen}(+5.27\%)} \\ 
    \textbf{RelaxQAT (ours)} & & \textbf{52.95} & \textbf{30.03} & \textbf{40.90} & 53.28 & \underline{64.09} & \textbf{48.25} & \textbf{40.14} & \textbf{52.73} & \textbf{47.04}  {\color{ForestGreen}\textbf{(+7.70\%)}} \\ 
    \hline 
    \end{tabular} } 
    \vspace{-3mm} 
    \caption{Evaluation of Qwen3-0.6B on Commonsense Reasoning, MMLU, and IF-Eval benchmarks. WG and HS denote WinoGrande and HellaSwag, respectively. ($\uparrow$): higher is better. {\color{ForestGreen} Green} and {\color{red} red} denote improvement and degradation relative to ParetoQ, respectively. \textbf{Bold} is best and \underline{underline} is second-best among QAT methods.} \label{tab:q30p6bllm} 
\end{table*}
\begin{table*}[t] 
    \centering 
    \scalebox{0.92}{ \setlength{\tabcolsep}{6pt} 
    \renewcommand{\arraystretch}{1.2} 
    \fontsize{8.0pt}{8.5pt}\selectfont 
    \begin{tabular}{l|c||ccccc||c|c|c||c} 
    \hline 
    \textbf{Method} & \textbf{Type} & \textbf{ArcE($\uparrow$)} & \textbf{ArcC($\uparrow$)} & \textbf{HS($\uparrow$)} & \textbf{WG($\uparrow$)} & \textbf{PIQA($\uparrow$)} & \textbf{CSR($\uparrow$)} & \textbf{MMLU($\uparrow$)} & \textbf{IFEval($\uparrow$)} & \textbf{Global Avg.($\uparrow$)} \\ 
    \hline 
    \noalign{\vskip 0.5pt} 
    \hline 
    FP & - & 75.63 & 54.35 & 73.01 & 69.69 & 77.97 & 70.13 & 69.70 & 85.31 & 75.04 \\ 
    \hline 
    AdaScale~\cite{adascaleptq} & PTQ & 24.32 & 25 & 26.44 & 51.22 & 51.79 & 35.75 & 24.16 & 16.12 & 25.34 \\ 
    \hline 
    ParetoQ~\cite{paretoq} & \multirow{6}{*}{QAT} & 73.94 & \underline{49.57} & \textbf{67.48} & 67.48 & 75.40 & 66.77 & 63.95 & 74.32 & 68.34 {\color{ForestGreen}(+0.00\%)} \\ 
    UPQ~\cite{upq} & & 71.84 & 46.67 & 67.10 & \underline{67.64} & \textbf{76.33} & 65.91 & 63.83 & 75.15 & 68.29 {\color{red}(-0.05\%)} \\
    LLM-QAT~\cite{llmqat} & & 73.95 & 47.78 & 67.25 & 65.90 & 75.68 & 66.11 & 64.03 & 73.95 & 68.03 {\color{red}(-0.31\%)} \\ 
    BitNet~\cite{bitnet158hf} & & 72.90 & 48.21 & 66.65 & 67.32 & 75.24 & 66.06 & 63.93 & \underline{76.29} & 68.76 {\color{ForestGreen}(+0.42\%)} \\ 
    WinQ~\cite{winq} & & \textbf{76.73} & \textbf{51.96} & 67.30 & 66.69 & 75.46 & \textbf{67.62} & \underline{64.44} & 75.88 & \underline{69.31} {\color{ForestGreen}(+0.97\%)} \\ 
    \textbf{RelaxQAT (ours)} & & \underline{74.41} & 48.89 & \underline{67.47} & \textbf{68.03} & \underline{75.79} & \underline{66.91} & \textbf{65.21} & \textbf{77.05} & \textbf{69.72} {\color{ForestGreen}\textbf{(+1.38\%)}} \\ \hline 
    \end{tabular} } 
    \vspace{-3mm} 
    \caption{Evaluation of Qwen3.5-4B on Commonsense Reasoning, MMLU, and IF-Eval benchmarks. WG and HS denote WinoGrande and HellaSwag, respectively. ($\uparrow$): higher is better. {\color{ForestGreen} Green} and {\color{red} red} denote improvement and degradation relative to ParetoQ, respectively. \textbf{Bold} is best and \underline{underline} is second-best among QAT methods.} \label{tab:qwn354bllm} 
\end{table*}
\begin{table*}[t] 
    \centering 
    \scalebox{0.92}{ \setlength{\tabcolsep}{6pt} 
    \renewcommand{\arraystretch}{1.2} 
    \fontsize{8.0pt}{8.5pt}\selectfont 
    \begin{tabular}{l|c||ccccc||c|c|c||c} 
    \hline 
    \textbf{Method} & \textbf{Type} & \textbf{ArcE($\uparrow$)} & \textbf{ArcC($\uparrow$)} & \textbf{HS($\uparrow$)} & \textbf{WG($\uparrow$)} & \textbf{PIQA($\uparrow$)} & \textbf{CSR($\uparrow$)} & \textbf{MMLU($\uparrow$)} & \textbf{IFEval($\uparrow$)} & \textbf{Global Avg.($\uparrow$)} \\ \hline \noalign{\vskip 0.5pt} 
    \hline 
    FP & - & 66.03 & 41.46 & 62.06 & 62.58 & 72.52 & 60.93 & 57.19 & 72.84 & 63.65 \\ 
    \hline 
    AdaScale~\cite{adascaleptq} & PTQ & 24.32 & 25.00 & 26.44 & 51.22 & 51.79 & 35.75 & 24.16 & 16.12 & 25.34 \\ 
    \hline 
    ParetoQ~\cite{paretoq} & \multirow{6}{*}{QAT} & 60.81 & \underline{37.28} & \textbf{57.38} & \underline{62.50} & \textbf{71.81} & \textbf{57.95} & \underline{54.42} & 62.12 & 58.16 {\color{ForestGreen}(+0.00\%)} \\ 
    UPQ~\cite{upq} & & 60.77 & 36.60 & \underline{56.92} & \textbf{62.66} & 71.27 & 57.64 & \textbf{54.63} & \underline{62.59} & \underline{58.28} {\color{ForestGreen}(+0.12\%)} \\ 
    LLM-QAT~\cite{llmqat} & & 58.54 & 36.77 & 56.62 & 62.11 & 71.16 & 57.04 & 53.40 & 62.15 & 57.53 {\color{red}(-0.63\%)} \\ 
    BitNet~\cite{bitnet158hf} & & 51.64 & 33.95 & 54.76 & 60.77 & 70.02 & 54.22 & 50.91 & 59.91 & 55.01 {\color{red}(-3.15\%)} \\
    WinQ~\cite{winq} & & \underline{61.36} & 36.94 & 56.70 & 62.03 & 71.32 & 57.67 & 53.43 & 59.07 & 56.72 {\color{red}(-1.44\%)} \\ 
    \textbf{RelaxQAT (ours)} & & \textbf{62.58} & \textbf{37.45} & 56.72 & 60.85 & \underline{71.59} & \underline{57.83} & 54.34 & \textbf{63.39} & \textbf{58.52} {\color{ForestGreen}\textbf{(+0.36\%)}} \\ 
    \hline \end{tabular} } 
    \vspace{-3mm} 
    \caption{Evaluation of Qwen3.5-2B on Commonsense Reasoning, MMLU, and IF-Eval benchmarks. WG and HS denote WinoGrande and HellaSwag, respectively. ($\uparrow$): higher is better. {\color{ForestGreen} Green} and {\color{red} red} denote improvement and degradation relative to ParetoQ, respectively. \textbf{Bold} is best and \underline{underline} is second-best among QAT methods.} \label{tab:qwn352bllm} 
\end{table*}

We now analyze the performance of RelaxQAT against several quantization baselines. Table~\ref{tab:q34bllm} presents the accuracy of various quantized 2-bit \texttt{Qwen3-4B-Instruct-2507} models on (\textit{i})~five common sense reasoning datasets, (\textit{ii})~their average (CSR), (\textit{iii})~MMLU, (\textit{iv})~IFEval, and (\textit{iv})~the global average of CSR, MMLU, and IFEval. As shown, AdaScale~\cite{adascaleptq}, a PTQ technique that combines leading methods like OmniQuant~\cite{shao2024omniquant} and FlexRound~\cite{lee2023flexround}, performs poorly in a 2-bit setting. In terms of QAT methods, ParetoQ achieves a global average accuracy of 69.03\%. Compared to this, RelaxQAT achieves 70.03\% global average accuracy ($\bm{+1\%}$ over ParetoQ) and also achieves the best overall accuracy on 7/9 columns reported in Table~\ref{tab:q34bllm} (with the second-best accuracy on the remaining two). The runner-up for the \texttt{Qwen3-4B-Instruct-2507} model class in global average is WinQ at 69.40\%. We beat WinQ by 0.63\%.

Next, we consider the \texttt{Qwen3-0.6B} to see if RelaxQAT is still effective on a smaller model. All benchmark evaluation results are shown in Table~\ref{tab:q30p6bllm}. Compared to ParetoQ's 39.34\% global average accuracy, we achieve a substantially higher 47.04\% ($\bm{+7.7\%}$) on this model class. The second-best method in Table~\ref{tab:q30p6bllm} is BitNet at 45.04\%. We outperform BitNet by $\bm{2\%}$. Thus, RelaxQAT outperforms even the second-best baseline by a significant margin for \texttt{Qwen3-0.6B}.

Let us now move on to a more recent model class: \texttt{Qwen3.5-4B}. Once again, Table~\ref{tab:qwn354bllm} shows a significant improvement of $\bm{1.38\%}$ accuracy over ParetoQ. The second best is WinQ and we outperform it by $\bm{0.41\%}$ on average. RelaxQAT also achieves the highest or second highest scores on 8 of the 9 columns reported in Table~\ref{tab:qwn354bllm}. Next, we also consider the \texttt{Qwen3.5-2B} model and show its results in Table~\ref{tab:qwn352bllm}. As evident, RelaxQAT outperform ParetoQ by $\bm{0.36\%}$ and achieves the best or second-best scores on 6 out of 9 columns reported. The second-best method for \texttt{Qwen3.5-2B} is UPQ with only $0.24\%$ lower accuracy than RelaxQAT. UPQ achieves highest or second highest scores on 5 out of 9 columns in Table~\ref{tab:qwn352bllm}. 


\subsection{Vision-Language Modeling Results}\label{sec:expvlms}
We now present the results on vision-language problems such as MMMU, DocVQA, and ScienceQA. Table~\ref{tab:qwn354bvlm} shows the results for \texttt{Qwen3.5-4B}. Clearly, we achieve the highest or second-highest accuracy on 3 of the 4 columns. Overall, RelaxQAT achieves the second highest average accuracy which is $\bm{0.82\%}$ above ParetoQ, but only 0.16\% below UPQ, the best performing method for this model. 
\begin{table}[t] 
    \centering 
    \scalebox{0.82}{ \setlength{\tabcolsep}{4pt} 
    \renewcommand{\arraystretch}{1.2} 
    \fontsize{8.0pt}{8.5pt}\selectfont 
    \begin{tabular}{l|c||ccc||c} 
    \hline 
    \textbf{Method} & \textbf{Type} & \textbf{MMMU($\uparrow$)} & \textbf{DocVQA($\uparrow$)} & \textbf{ScienceQA($\uparrow$)} & \textbf{Avg.($\uparrow$)} \\ 
    \hline 
    \noalign{\vskip 0.5pt} 
    \hline 
    FP & - & 41.22 & 94.51 & 85.52 & 73.75 \\ 
    \hline
    AdaScale& PTQ & 26.00 & 0 & 0 & 8.66 \\ 
    \hline 
    ParetoQ& \multirow{6}{*}{QAT} & \underline{40.78} & 90.42 & 81.85 & 71.02 {\color{ForestGreen}(+0.00\%)} \\
    UPQ& & \textbf{42.44} & \underline{90.80} & \underline{82.75} & \textbf{72.00} {\color{ForestGreen}\textbf{(+0.98\%)}} \\ LLM-QAT & & 39.22 & 90.76 & 79.67 & 69.88 {\color{red}(-1.14\%)} \\
    BitNet& & 40.33 & 90.60 & 81.71 & 70.88 {\color{red}(-0.14\%)} \\ 
    WinQ& & 34.00 & 7.76$^*$ & 65.94 & 35.90 {\color{red}(-35.12\%)} \\ 
    \textbf{RelaxQAT} & & 40.33 & \textbf{91.25} & \textbf{83.94} & \underline{71.84} {\color{ForestGreen}(+0.82\%)} \\ 
    \hline 
    \end{tabular} } 
    \vspace{-1mm} 
    \caption{Evaluation of Qwen3.5-4B on MMMU, DocVQA, and ScienceQA. ($\uparrow$): higher is better. {\color{ForestGreen} Green} and {\color{red} red} denote improvement and degradation relative to ParetoQ, respectively. \textbf{Bold} is best and \underline{underline} is second-best among QAT methods. $^*$WinQ showed some training instability which caused loss of accuracy.} \label{tab:qwn354bvlm} 
\end{table}
\begin{table}[t] 
    \centering 
    \scalebox{0.82}{ \setlength{\tabcolsep}{4pt} 
    \renewcommand{\arraystretch}{1.2} 
    \fontsize{8.0pt}{8.5pt}\selectfont 
    \begin{tabular}{l|c||ccc||c} 
    \hline 
    \textbf{Method} & \textbf{Type} & \textbf{MMMU($\uparrow$)} & \textbf{DocVQA($\uparrow$)} & \textbf{ScienceQA($\uparrow$)} & \textbf{Avg.($\uparrow$)} \\ 
    \hline 
    \noalign{\vskip 0.5pt} 
    \hline 
    FP & - & 39.77 & 92.05 & 82.89 & 71.57 \\ \hline AdaScale & PTQ & 25.55 & 0 & 0 & 8.52 \\ 
    \hline
    ParetoQ & \multirow{6}{*}{QAT} & \underline{31.77} & 87.73 & 52.30 & 57.26 {\color{ForestGreen}(+0.00\%)} \\ 
    UPQ & & 30.55 & \textbf{88.09} & \underline{60.68} & \underline{59.77} {\color{ForestGreen}(+2.51\%)} \\ 
    LLM-QAT & & 30.66 & 87.34 & 56.66 & 58.22 {\color{ForestGreen}(+0.96\%)} \\ 
    BitNet & & 29.11 & \underline{87.90} & 59.29 & 58.76 {\color{ForestGreen}(+1.50\%)} \\ 
    WinQ & & 29.11 & 72.16 & 57.01 & 52.76 {\color{red}(-4.50\%)} \\ 
    \textbf{RelaxQAT} & & \textbf{33.11} & 87.71 & \textbf{69.75} & \textbf{63.52} {\color{ForestGreen}\textbf{(+6.26\%)}} \\ 
    \hline 
    \end{tabular} } 
    \vspace{-1mm} 
    \caption{Evaluation of Qwen3.5-2B on MMMU, DocVQA, and ScienceQA. ($\uparrow$): higher is better. {\color{ForestGreen} Green} and {\color{red} red} denote improvement and degradation relative to ParetoQ, respectively. \textbf{Bold} is best and \underline{underline} is second-best among QAT methods.} \label{tab:qwn352bvlm} 
\end{table}
\begin{table*}[h] 
    \centering 
    \scalebox{1.0}{ \setlength{\tabcolsep}{6pt} 
    \renewcommand{\arraystretch}{1.2} 
    \fontsize{8.0pt}{8.5pt}\selectfont 
    \begin{tabular}{l||ccccc||c|c|c||c} 
    \hline 
    \textbf{$\bm{m_s}$} & \textbf{ArcE($\uparrow$)} & \textbf{ArcC($\uparrow$)} & \textbf{HS($\uparrow$)} & \textbf{WG($\uparrow$)} & \textbf{PIQA($\uparrow$)} & \textbf{CSR($\uparrow$)} & \textbf{MMLU($\uparrow$)} & \textbf{IFEval($\uparrow$)} & \textbf{Global Avg.($\uparrow$)} \\ 
    \hline\hline
    \noalign{\vskip 0.5pt} 
    FP & 56.01 & 34.55 & 47.29 & 55.48 & 67.73 & 52.21 & 47.25 & 64.78 & 54.74 \\ 
    \hline 
    ParetoQ & 50.79 & 27.64 & 40.19 & 52.40 & 64.03 & 47.01 & 37.18 & 33.85 & 39.34 {\color{ForestGreen}(+0.00\%)} \\ 
    \hline 
     400 & \underline{53.11} & 29.01 & 40.82 & 53.91 & 63.82 & 48.13 & 40.01 & \underline{51.28} & \underline{46.47} {\color{ForestGreen}(+7.13\%)} \\ 
    800 & 52.95 & \textbf{30.03} & 40.90 & 53.28 & \textbf{64.09} & 48.25 & \underline{40.14} & \textbf{52.73} & \textbf{47.04} \textbf{{\color{ForestGreen}(+7.70\%)}} \\ 
    1200 & \textbf{53.24} & \underline{29.10} & 40.88 & \underline{55.49} & \underline{63.98} & \textbf{48.54} & 40.09 & 49.78 & 46.14 {\color{ForestGreen}(+6.80\%)}  \\ 
    1600 & 50.88 & 28.92 & \underline{41.19} & 54.14 & 63.71 & 47.77 & \textbf{40.57} & 49.26 & 45.87 {\color{ForestGreen}(+6.53\%)} \\
    2000 & 51.14 & 28.75 & \textbf{41.22} & \textbf{56.27} & \underline{63.98} & \underline{48.27} & 40.12 & 48.99 & 45.79 {\color{ForestGreen}(+6.45\%)} \\  
    \hline 
    \end{tabular} } 
    \vspace{-1mm} 
    \caption{Ablation study of Qwen3-0.6B with varying maximum schedule steps ($m_s$) and number of full cycles ($n$) fixed to 1 (after the initial half cycle of bit schedule). Evaluation is performed on Commonsense Reasoning, MMLU, and IF-Eval benchmarks. WG and HS denote WinoGrande and HellaSwag, respectively. ($\uparrow$): higher is better. \textbf{Bold} is best and \underline{underline} is second-best among QAT methods.} \label{tab:abl1} 
\end{table*}
\begin{table*}[h] 
    \centering 
    \scalebox{1.0}{ \setlength{\tabcolsep}{6pt} 
    \renewcommand{\arraystretch}{1.2} 
    \fontsize{8.0pt}{8.5pt}\selectfont 
    \begin{tabular}{l||ccccc||c|c|c||c} 
    \hline 
    \textbf{$\bm{n}$} & \textbf{ArcE($\uparrow$)} & \textbf{ArcC($\uparrow$)} & \textbf{HS($\uparrow$)} & \textbf{WG($\uparrow$)} & \textbf{PIQA($\uparrow$)} & \textbf{CSR($\uparrow$)} & \textbf{MMLU($\uparrow$)} & \textbf{IFEval($\uparrow$)} & \textbf{Global Avg.($\uparrow$)} \\ 
    \hline\hline
    \noalign{\vskip 0.5pt} 
    FP & 56.01 & 34.55 & 47.29 & 55.48 & 67.73 & 52.21 & 47.25 & 64.78 & 54.74 \\ 
    \hline 
    ParetoQ & 50.79 & 27.64 & 40.19 & 52.40 & 64.03 & 47.01 & 37.18 & 33.85 & 39.34 {\color{ForestGreen}(+0.00\%)} \\ 
    \hline
     0 & \underline{51.68} & \textbf{29.10} & 41.01 & 54.14 & 63.22 & 47.83 & 39.84 & 43.81 & 43.83 {\color{ForestGreen}(+4.49\%)} \\
    1 & 49.66 & 28.75 & 40.78 & 53.43 & \underline{63.98} & 47.32 & 39.45 & \textbf{52.02} & \textbf{46.26} \textbf{{\color{ForestGreen}(+6.92\%)}} \\
    2 & \textbf{52.95} & \underline{29.01} & 40.68 & 53.91 & 63.38 & \underline{47.99} & \textbf{40.67} & 49.61 & \underline{46.09} {\color{ForestGreen}(+6.75\%)}  \\
    3 & 51.18 & 28.41 & \textbf{41.22} & \textbf{55.01} & \textbf{64.42} & \textbf{48.05} & 39.39 & \underline{49.71} & 45.72 {\color{ForestGreen}(+6.38\%)} \\
    4 & 50.80 & 28.84 & \underline{41.09} & \underline{54.78} & 63.33 & 47.77 & \underline{40.45} & 48.44 & 45.55 {\color{ForestGreen}(+6.21\%)} \\  
    \hline 
    \end{tabular} } 
    \vspace{-1mm} 
    \caption{Ablation study of Qwen3-0.6B with varying number of full cycles, $n$ (after the initial half cycle of bit schedule), with maximum schedule steps ($m_s$) fixed to 400. Evaluation is performed on Commonsense Reasoning, MMLU, and IF-Eval benchmarks. WG and HS denote WinoGrande and HellaSwag, respectively. ($\uparrow$): higher is better. \textbf{Bold} is best and \underline{underline} is second-best among QAT methods.} \label{tab:abl2} 
\end{table*}

For \texttt{Qwen3.5-2B}, the results are shown in Table~\ref{tab:qwn352bvlm}. As evident, RelaxQAT demonstrates strong VLM performance with a massive $\bm{6.26\%}$ accuracy gain over ParetoQ. We also see a nearly $\bm{9\%}$ gain in the ScienceQA score for RelaxQAT compared to the second-best method (UPQ). On average, RelaxQAT outperforms UPQ by a large ($\bm{3.75\%}$) margin. 

Our results demonstrate how RelaxQAT outperforms baselines across the board for both LLM and VLM tasks. \textit{Importantly, no other single method consistently scores the highest or second-highest for all these different tasks}; the second highest position is shared by WinQ, UPQ, and BitNet. This further highlights the effectiveness of RelaxQAT.

\subsection{Ablations}
In this section, we evaluate the impact of RelaxQAT hyperparameters such as maximum schedule steps $m_s$ and number of full cycles $n$. To this end, we conduct two ablation studies using the \texttt{Qwen3-0.6B} model. We also show whether RelaxQAT works on model classes other than the Qwen family.

\paragraph{Varying $\bm{m_s}$.} While fixing $n=1$ full cycle after the initial half cycle of the bit schedule, we vary $m_s$ between $\{400, 800, 1200, 1600, 2000\}$. The results are shown in Table~\ref{tab:abl1}. We make two key observations: (a)~RelaxQAT is fairly robust across different choices of maximum schedule steps. Specifically, all $m_s$ values significantly outperform ParetoQ (by $\bm{6.45\%}$ to $\bm{7.7\%}$) and also beat the second highest BitNet method shown in Table~\ref{tab:q30p6bllm}. (b)~The values of $m_s\in\{400, 800, 1200\}$ typically perform the best, which is what we have used throughout this work.

\paragraph{Varying $\bm{n}$.} Next, we change the number of full cycles $n\in\{0,1,2,3,4\}$ after the initial half cycle for $m_s=400$. Table~\ref{tab:abl2} shows that RelaxQAT is once again quite robust to the choice of $n$ since all values provide a substantial accuracy boost compared to ParetoQ ($\bm{4.5\%}$ to $\bm{7\%}$). Moreover, $n=1$ performs the best among all other $n$ values considered. Of note, RelaxQAT with $n=0$, i.e., just the initial half cycle of bit schedule and no full cycles, outperforms ParetoQ by about $\bm{4.5\%}$. Most interestingly, \textit{all} non-zero full cycle cases ($n\geq1$) outperform RelaxQAT with $n=0$ by a large margin of up to nearly $\bm{2.5\%}$. This demonstrates that while the initial half cycle dramatically speeds up the early convergence, further bit schedule relaxation cycles result in significant additional accuracy gains.

\paragraph{What about model classes outside the Qwen family?}
To address this, we present results for \texttt{Gemma-3-1B-it} model in Table~\ref{tab:g31bllm} in Appendix~\ref{app:ablations}. Once again, RelaxQAT significantly outperforms ParetoQ by $\bm{3.35\%}$ with highest or second highest scores in 6 out of 9 columns. Overall, RelaxQAT outperforms the second best method WinQ by $\bm{2.63\%}$, a significant margin. Most notably, RelaxQAT achieves a nearly $\bm{6\%}$ higher IFEval score than the second best method.

\subsection{Does RelaxQAT work for Ternary Quantization?}
Finally, we investigate whether RelaxQAT works for other quantization schemes such as ternary (1.58) bits: $\{-1, 0, 1\}$ integer grid. Table~\ref{tab:tq34bllm} in Appendix~\ref{app:ternaryquantization} shows the accuracy of ternary \texttt{Qwen3-4B-Instruct-2507} models. As evident, RelaxQAT performs the best with $\bm{4.2\%}$ higher accuracy than ParetoQ. Similar results are observed for Table~\ref{tab:tqwn352bllm} in Appendix~\ref{app:ternaryquantization} for the ternary \texttt{Qwen3.5-2B} model. RelaxQAT outperforms ParetoQ by $\bm{1.38\%}$. Hence, RelaxQAT is a versatile technique that strongly outperforms leading methods for both ternary (1.58-bit) and 2-bit quantization.

\section{Conclusion and Future Work}\label{sec::conc}
In this paper, we have explored a new paradigm which aims to create a continuous relaxation of the QAT process. To this end, we have proposed RelaxQAT which dynamically varies the bitwidth during training in a continuous floating point domain. RelaxQAT gradually alters the bit precision of model weights in order to accelerate the early training convergence, as well as to escape any flat regions in the low-bit quantization loss surfaces. We have also presented simple theoretical and empirical insights into why RelaxQAT helps QAT. Our extensive experiments  demonstrate accuracy gains of up to $\bm{6.3\%}$ on vision and up to $\bm{7.7\%}$ on natural language tasks, compared to the leading QAT methods. 

Finally, although we have considered weight only quantization in this work, RelaxQAT can also be applied to activation or KV-cache quantization. We leave this for future work.

\bibliography{references}

@article{upq,
  title={{Unifying Block-wise PTQ and Distillation-based QAT for Progressive Quantization toward 2-bit Instruction-Tuned LLMs}},
  author={Lee, Jung Hyun and Shin, Seungjae and Kim, Vinnam and You, Jaeseong and Chen, An},
  journal={arXiv preprint arXiv:2506.09104},
  year={2025}
}

@online{bitnet158hf,
  author = {{Hugging Face}},
  title = {Fine-tuning LLMs to 1.58bit: Extreme Quantization Made Easy},
  year = {2024},
  url = {https://huggingface.co/blog/1_58_llm_extreme_quantization},
  urldate = {2026-08-04}
}

@inproceedings{lee2023flexround,
  title={Flexround: Learnable rounding based on element-wise division for post-training quantization},
  author={Lee, Jung Hyun and Kim, Jeonghoon and Kwon, Se Jung and Lee, Dongsoo},
  booktitle={International Conference on Machine Learning},
  pages={18913--18939},
  year={2023},
  organization={PMLR}
}

@online{adascaleptq,
  author = {{AIMET}},
  title = {AdaScale PTQ},
  year = {2026},
  url = {https://quic.github.io/aimet-pages/releases/latest/ptq_techniques/adascale.html},
  urldate = {2026-09-15}
}

@article{paretoq,
  title={Paretoq: Improving scaling laws in extremely low-bit llm quantization},
  author={Liu, Zechun and Zhao, Changsheng and Huang, Hanxian and Chen, Sijia and Zhang, Jing and Zhao, Jiawei and Roy, Scott and Jin, Lisa and Xiong, Yunyang and Shi, Yangyang and others},
  journal={Advances in Neural Information Processing Systems},
  volume={38},
  pages={91311--91336},
  year={2025}
}

@InProceedings{winq,
  title = 	 {{W}in{Q}: Accelerating Quantization-Aware Training of Language Models Around Saddle Points},
  author =       {Li, Dongyue and Liu, Zechun and Yi, Kai and Zhang, Zhenshuo and Zhao, Changsheng and Krishnamoorthi, Raghuraman and Khaitan, Harshit and Zhang, Hongyang R. and Li, Steven},
  booktitle = 	 {Proceedings of the 43rd International Conference on Machine Learning},
  pages = 	 {68772--68794},
  year = 	 {2026},
  editor = 	 {Zhang, Tong and Dudik, Miroslav and Jaggi, Martin and Agarwal, Alekh and Li, Sharon and Schuurmans, Dale and Zhu, Jerry and Berkenkamp, Felix and Dong, Hanze and Bietti, Alberto},
  volume = 	 {306},
  series = 	 {Proceedings of Machine Learning Research},
  month = 	 {06--11 Jul},
  publisher =    {PMLR},
  url = 	 {https://proceedings.mlr.press/v306/li26dr.html}
}

@article{lsq,
  title={Learned step size quantization},
  author={Esser, Steven K and McKinstry, Jeffrey L and Bablani, Deepika and Appuswamy, Rathinakumar and Modha, Dharmendra S},
  journal={arXiv preprint arXiv:1902.08153},
  year={2019}
}

@article{gong2025survey,
  title={A survey of low-bit large language models: Basics, systems, and algorithms},
  author={Gong, Ruihao and Ding, Yifu and Wang, Zining and Lv, Chengtao and Zheng, Xingyu and Du, Jinyang and Yong, Yang and Gu, Shiqiao and Qin, Haotong and Guo, Jinyang and others},
  journal={Neural networks},
  volume={192},
  pages={107856},
  year={2025},
  publisher={Elsevier}
}

@article{nagel2021white,
  title={A white paper on neural network quantization},
  author={Nagel, Markus and Fournarakis, Marios and Amjad, Rana Ali and Bondarenko, Yelysei and Van Baalen, Mart and Blankevoort, Tijmen},
  journal={arXiv preprint arXiv:2106.08295},
  year={2021}
}

@article{song2026target,
  title={A Target-Centric Survey of Quantization-Aware Training},
  author={Song, Jiamin and Zhao, Mengjie and Wang, Zijing and Liu, Yongkang and Li, Qian and Feng, Shi and Ren, Feiliang and Wang, Daling and Sch{\"u}tze, Hinrich},
  journal={arXiv preprint arXiv:2608.29667},
  year={2026}
}

@article{lin2024awq,
  title={Awq: Activation-aware weight quantization for on-device llm compression and acceleration},
  author={Lin, Ji and Tang, Jiaming and Tang, Haotian and Yang, Shang and Chen, Wei-Ming and Wang, Wei-Chen and Xiao, Guangxuan and Dang, Xingyu and Gan, Chuang and Han, Song},
  journal={Proceedings of machine learning and systems},
  volume={6},
  pages={87--100},
  year={2024}
}

@inproceedings{liu2025spinquant,
  title={Spinquant: Llm quantization with learned rotations},
  author={Liu, Zechun and Zhao, Changsheng and Fedorov, Igor and Soran, Bilge and Choudhary, Dhruv and Krishnamoorthi, Raghuraman and Chandra, Vikas and Tian, Yuandong and Blankevoort, Tijmen},
  booktitle={International Conference on Learning Representations},
  volume={2025},
  pages={92009--92032},
  year={2025}
}

@article{sun2020ultra,
  title={Ultra-low precision 4-bit training of deep neural networks},
  author={Sun, Xiao and Wang, Naigang and Chen, Chia-Yu and Ni, Jiamin and Agrawal, Ankur and Cui, Xiaodong and Venkataramani, Swagath and El Maghraoui, Kaoutar and Srinivasan, Vijayalakshmi Viji and Gopalakrishnan, Kailash},
  journal={Advances in Neural Information Processing Systems},
  volume={33},
  pages={1796--1807},
  year={2020}
}

@article{frantar2022gptq,
  title={Gptq: Accurate post-training quantization for generative pre-trained transformers},
  author={Frantar, Elias and Ashkboos, Saleh and Hoefler, Torsten and Alistarh, Dan},
  journal={arXiv preprint arXiv:2210.17323},
  year={2022}
}

@article{yi2026jacquant,
  title={JacQuant: STE-Free Quantization-Aware Training via Learned Jacobian Surrogates},
  author={Yi, Kai and Vivekraja, Vignesh and Khaitan, Harshit and Li, Steven},
  journal={arXiv preprint arXiv:2605.25469},
  year={2026}
}

@inproceedings{llmqat,
  title={Llm-qat: Data-free quantization aware training for large language models},
  author={Liu, Zechun and Oguz, Barlas and Zhao, Changsheng and Chang, Ernie and Stock, Pierre and Mehdad, Yashar and Shi, Yangyang and Krishnamoorthi, Raghuraman and Chandra, Vikas},
  booktitle={Findings of the association for computational linguistics: ACL 2024},
  pages={467--484},
  year={2024}
}

@article{ashkboos2024quarot,
  title={Quarot: Outlier-free 4-bit inference in rotated llms},
  author={Ashkboos, Saleh and Mohtashami, Amirkeivan and Croci, Maximilian L and Li, Bo and Cameron, Pashmina and Jaggi, Martin and Alistarh, Dan and Hoefler, Torsten and Hensman, James},
  journal={Advances in Neural Information Processing Systems},
  volume={37},
  pages={100213--100240},
  year={2024}
}

@inproceedings{chen2025efficientqat,
  title={Efficientqat: Efficient quantization-aware training for large language models},
  author={Chen, Mengzhao and Shao, Wenqi and Xu, Peng and Wang, Jiahao and Gao, Peng and Zhang, Kaipeng and Luo, Ping},
  booktitle={Proceedings of the 63rd Annual Meeting of the Association for Computational Linguistics (Volume 1: Long Papers)},
  pages={10081--10100},
  year={2025}
}

@inproceedings{tillet2019triton,
  title={Triton: an intermediate language and compiler for tiled neural network computations},
  author={Tillet, Philippe and Kung, Hsiang-Tsung and Cox, David},
  booktitle={Proceedings of the 3rd ACM SIGPLAN International Workshop on Machine Learning and Programming Languages},
  pages={10--19},
  year={2019}
}

@article{qwen3,
  title={Qwen3 technical report},
  author={Yang, An and Li, Anfeng and Yang, Baosong and Zhang, Beichen and Hui, Binyuan and Zheng, Bo and Yu, Bowen and Gao, Chang and Huang, Chengen and Lv, Chenxu and others},
  journal={arXiv preprint arXiv:2505.09388},
  year={2025}
}

@article{choi2018pact,
  title={Pact: Parameterized clipping activation for quantized neural networks},
  author={Choi, Jungwook and Wang, Zhuo and Venkataramani, Swagath and Chuang, Pierce I-Jen and Srinivasan, Vijayalakshmi and Gopalakrishnan, Kailash},
  journal={arXiv preprint arXiv:1805.06085},
  year={2018}
}

@inproceedings{bhalgat2020lsq+,
  title={Lsq+: Improving low-bit quantization through learnable offsets and better initialization},
  author={Bhalgat, Yash and Lee, Jinwon and Nagel, Markus and Blankevoort, Tijmen and Kwak, Nojun},
  booktitle={2020 IEEE/CVF Conference on Computer Vision and Pattern Recognition Workshops (CVPRW)},
  pages={2978--2985},
  year={2020},
  organization={IEEE}
}

@inproceedings{dettmers2023case,
  title={The case for 4-bit precision: k-bit inference scaling laws},
  author={Dettmers, Tim and Zettlemoyer, Luke},
  booktitle={International Conference on Machine Learning},
  pages={7750--7774},
  year={2023},
  organization={PMLR}
}

@inproceedings{vsubic2026balancing,
  title={Balancing Compute in LLM Inference: Model Selection, Quantization, and Test-Time Scaling},
  author={{\v{S}}ubi{\'c}, Tomislav and Cozzini, Stefano},
  booktitle={Proceedings of the Sixth European Workshop on Machine Learning and Systems},
  pages={433--438},
  year={2026}
}

@article{wang2024bitnet,
  title={Bitnet a4. 8: 4-bit activations for 1-bit llms},
  author={Wang, Hongyu and Ma, Shuming and Wei, Furu},
  journal={arXiv preprint arXiv:2411.04965},
  year={2024}
}

@article{wang2025bitnetLLM,
  title={Bitnet: 1-bit pre-training for large language models},
  author={Wang, Hongyu and Ma, Shuming and Ma, Lingxiao and Wang, Lei and Wang, Wenhui and Dong, Li and Huang, Shaohan and Wang, Huaijie and Xue, Jilong and Wang, Ruiping and others},
  journal={Journal of Machine Learning Research},
  volume={26},
  number={125},
  pages={1--29},
  year={2025}
}

@inproceedings{wang2025bitnetedge,
  title={Bitnet. cpp: Efficient edge inference for ternary llms},
  author={Wang, Jinheng and Zhou, Hansong and Song, Ting and Cao, Shijie and Xia, Yan and Cao, Ting and Wei, Jianyu and Ma, Shuming and Wang, Hongyu and Wei, Furu},
  booktitle={Proceedings of the 63rd Annual Meeting of the Association for Computational Linguistics (Volume 1: Long Papers)},
  pages={9305--9322},
  year={2025}
}

@article{lv2026makes,
  title={What makes low-bit quantization-aware training work for reasoning llms? a systematic study},
  author={Lv, Keyu and Zhang, Manyi and Xia, Xiaobo and Ni, Jingchen and Yan, Shannan and Yu, Xianzhi and Hou, Lu and Yuan, Chun and Bai, Haoli},
  journal={arXiv preprint arXiv:2601.14888},
  year={2026}
}

@inproceedings{xu2026bit,
  title={Bit-by-bit: Progressive qat strategy with outlier channel splitting for stable low-bit llms},
  author={Xu, Binxing and Gu, Hao and Li, Lujun and Wang, Hao and Liu, Bei and Liu, Jiacheng and Zhu, Qiyuan and Yang, Xintong and Li, Chao and Han, Sirui and others},
  booktitle={Proceedings of the 64th Annual Meeting of the Association for Computational Linguistics (Volume 1: Long Papers)},
  pages={22284--22299},
  year={2026}
}

@article{wei2022qdrop,
  title={Qdrop: Randomly dropping quantization for extremely low-bit post-training quantization},
  author={Wei, Xiuying and Gong, Ruihao and Li, Yuhang and Liu, Xianglong and Yu, Fengwei},
  journal={arXiv preprint arXiv:2203.05740},
  year={2022}
}

@inproceedings{shao2024omniquant,
  title={Omniquant: Omnidirectionally calibrated quantization for large language models},
  author={Shao, Wenqi and Chen, Mengzhao and Zhang, Zhaoyang and Xu, Peng and Zhao, Lirui and Li, Zhiqian and Zhang, Kaipeng and Peng, Gao and Qiao, Yu and Luo, Ping},
  booktitle={International Conference on Learning Representations},
  volume={2024},
  pages={45472--45496},
  year={2024}
}

@inproceedings{lee2024owq,
  title={Owq: Outlier-aware weight quantization for efficient fine-tuning and inference of large language models},
  author={Lee, Changhun and Jin, Jungyu and Kim, Taesu and Kim, Hyungjun and Park, Eunhyeok},
  booktitle={Proceedings of the AAAI Conference on Artificial Intelligence},
  volume={38},
  number={12},
  pages={13355--13364},
  year={2024}
}

@inproceedings{wang2025outlier,
  title={Outlier-aware post-training quantization for image super-resolution},
  author={Wang, Hailing and Lu, Jianglin and Zhang, Yitian and Fu, Yun},
  booktitle={2025 IEEE/CVF International Conference on Computer Vision (ICCV)},
  pages={16175--16184},
  year={2025},
  organization={IEEE}
}

@article{wang2025ofq,
  title={Ofq-llm: Outlier-flexing quantization for efficient low-bit large language model acceleration},
  author={Wang, Gang and Cai, Siqi and Li, Wenjie and Lyu, Dongxu and He, Guanghui},
  journal={IEEE Transactions on Circuits and Systems I: Regular Papers},
  volume={72},
  number={8},
  pages={4077--4090},
  year={2025},
  publisher={IEEE}
}

@article{xu2025butterflyquant,
  title={Butterflyquant: Ultra-low-bit llm quantization through learnable orthogonal butterfly transforms},
  author={Xu, Bingxin and Dong, Zhen and Elachqar, Oussama and Shang, Yuzhang},
  journal={arXiv preprint arXiv:2509.09679},
  year={2025}
}

@misc{allal2025smollm2smolgoesbig,
      title={SmolLM2: When Smol Goes Big -- Data-Centric Training of a Small Language Model}, 
      author={Loubna Ben Allal and Anton Lozhkov and Elie Bakouch and Gabriel Martín Blázquez and Guilherme Penedo and Lewis Tunstall and Andrés Marafioti and Hynek Kydlíček and Agustín Piqueres Lajarín and Vaibhav Srivastav and Joshua Lochner and Caleb Fahlgren and Xuan-Son Nguyen and Clémentine Fourrier and Ben Burtenshaw and Hugo Larcher and Haojun Zhao and Cyril Zakka and Mathieu Morlon and Colin Raffel and Leandro von Werra and Thomas Wolf},
      year={2025},
      eprint={2502.02737},
      archivePrefix={arXiv},
      primaryClass={cs.CL},
      url={https://arxiv.org/abs/2502.02737}, 
}

@article{liu2024apigen,
  title={APIGen: Automated Pipeline for Generating Verifiable and Diverse Function-Calling Datasets},
  author={Liu, Zuxin and Hoang, Thai and Zhang, Jianguo and Zhu, Ming and Lan, Tian and Kokane, Shirley and Tan, Juntao and Yao, Weiran and Liu, Zhiwei and Feng, Yihao and others},
  journal={arXiv preprint arXiv:2406.18518},
  year={2024}
}

@online{tulu3sftolmo2mixture,
  author = {{Allen Institute for AI}},
  title = {Tulu-3-SFT-OLMo-2-Mixture},
  url = {https://huggingface.co/datasets/allenai/tulu-3-sft-olmo-2-mixture},
  urldate = {2026-10-29}
}

@online{llamanemotronvlmdataset,
  author = {{NVIDIA}},
  title = {Llama-Nemotron-VLM-Dataset-v1},
  url = {https://huggingface.co/datasets/nvidia/Llama-Nemotron-VLM-Dataset-v1},
  urldate = {2026-10-29}
}

@article{clark2018think,
  title={Think you have solved question answering? try arc, the ai2 reasoning challenge},
  author={Clark, Peter and Cowhey, Isaac and Etzioni, Oren and Khot, Tushar and Sabharwal, Ashish and Schoenick, Carissa and Tafjord, Oyvind},
  journal={arXiv preprint arXiv:1803.05457},
  year={2018}
}

@inproceedings{bisk2020piqa,
  title={Piqa: Reasoning about physical commonsense in natural language},
  author={Bisk, Yonatan and Zellers, Rowan and Gao, Jianfeng and Choi, Yejin and others},
  booktitle={Proceedings of the AAAI conference on artificial intelligence},
  volume={34},
  number={05},
  pages={7432--7439},
  year={2020}
}

@inproceedings{zellers2019hellaswag,
  title={Hellaswag: Can a machine really finish your sentence?},
  author={Zellers, Rowan and Holtzman, Ari and Bisk, Yonatan and Farhadi, Ali and Choi, Yejin},
  booktitle={Proceedings of the 57th annual meeting of the association for computational linguistics},
  pages={4791--4800},
  year={2019}
}

@article{sakaguchi2021winogrande,
  title={Winogrande: An adversarial winograd schema challenge at scale},
  author={Sakaguchi, Keisuke and Bras, Ronan Le and Bhagavatula, Chandra and Choi, Yejin},
  journal={Communications of the ACM},
  volume={64},
  number={9},
  pages={99--106},
  year={2021},
  publisher={ACM, New York, NY, USA}
}

@article{hendrycks2020measuring,
  title={Measuring massive multitask language understanding},
  author={Hendrycks, Dan and Burns, Collin and Basart, Steven and Zou, Andy and Mazeika, Mantas and Song, Dawn and Steinhardt, Jacob},
  journal={arXiv preprint arXiv:2009.03300},
  year={2020}
}

@article{zhou2023instruction,
  title={Instruction-following evaluation for large language models},
  author={Zhou, Jeffrey and Lu, Tianjian and Mishra, Swaroop and Brahma, Siddhartha and Basu, Sujoy and Luan, Yi and Zhou, Denny and Hou, Le},
  journal={arXiv preprint arXiv:2311.07911},
  year={2023}
}

@inproceedings{yue2024mmmu,
  title={Mmmu: A massive multi-discipline multimodal understanding and reasoning benchmark for expert agi},
  author={Yue, Xiang and Ni, Yuansheng and Zhang, Kai and Zheng, Tianyu and Liu, Ruoqi and Zhang, Ge and Stevens, Samuel and Jiang, Dongfu and Ren, Weiming and Sun, Yuxuan and others},
  booktitle={Proceedings of the IEEE/CVF conference on computer vision and pattern recognition},
  pages={9556--9567},
  year={2024}
}

@inproceedings{mathew2021docvqa,
  title={Docvqa: A dataset for vqa on document images},
  author={Mathew, Minesh and Karatzas, Dimosthenis and Jawahar, CV},
  booktitle={2021 IEEE Winter Conference on Applications of Computer Vision (WACV)},
  pages={2199--2208},
  year={2021},
  organization={IEEE}
}

@article{lu2022learn,
  title={Learn to explain: Multimodal reasoning via thought chains for science question answering},
  author={Lu, Pan and Mishra, Swaroop and Xia, Tanglin and Qiu, Liang and Chang, Kai-Wei and Zhu, Song-Chun and Tafjord, Oyvind and Clark, Peter and Kalyan, Ashwin},
  journal={Advances in neural information processing systems},
  volume={35},
  pages={2507--2521},
  year={2022}
}
\bibliographystyle{mlsys2025}

\section{AI Usage Disclosure}
No AI was used in ideation, literature search, experiment design, code implementation, and writing of this paper. Claude code was only used to write a python code to process the Tensorboard logs of a few  trained models and to plot these training loss values in matplotlib (which are shown in Fig~\ref{fig:tl}). 

\clearpage
\appendix
\twocolumn[{\centering \large \textbf{Supplementary Information:} A Continuous Relaxation of Quantization-Aware Training\vspace{+10mm}}]

\section{Existing SEQ vs. RelaxSEQ}\label{app:extremeval}
Let us now look at some concrete loss and test accuracy behavior for the existing ParetoQ SEQ~\cite{paretoq} (Clip-then-Round) vs. the proposed RelaxSEQ (Round-then-Clip) when trained under a specific bitwidth schedule. As mentioned before, existing SEQ (due to Clip-then-Round) loses extreme integers as the schedule reduces bitwidth from $b_s$ to $b_e$ bits. In contrast, RelaxSEQ results in a smoothly changing extreme values. Our hypothesis is that the sudden change in extreme integers can cause the loss to spike. 

\begin{figure}[H]
  \centering
   \includegraphics[width=0.9\linewidth]{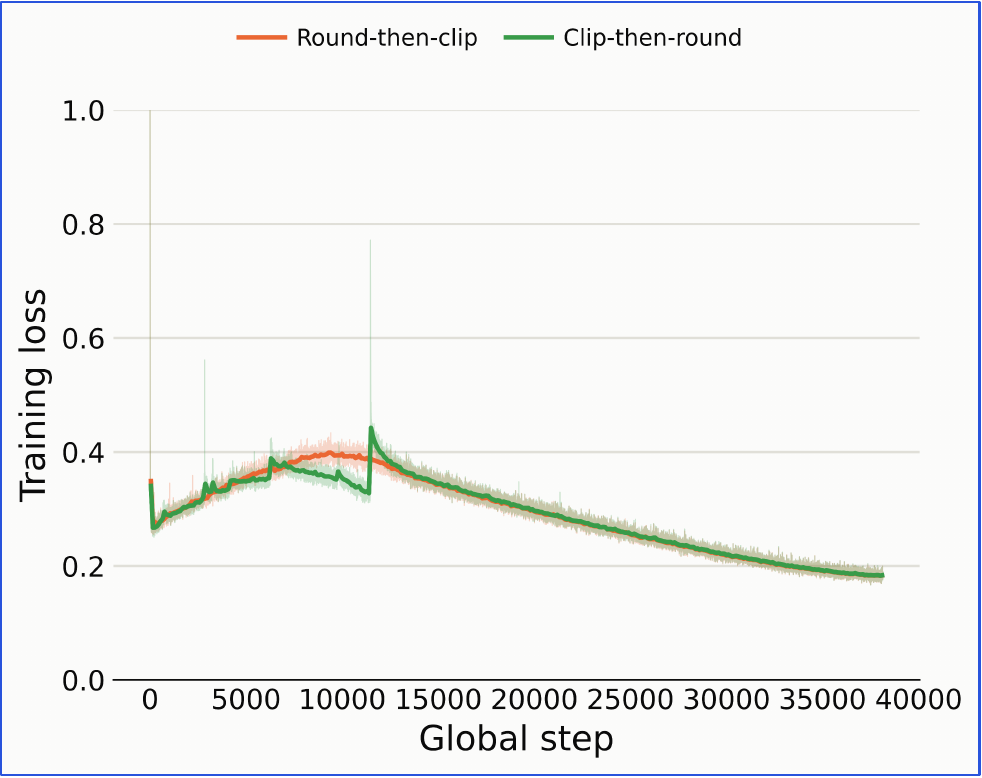}\vspace{-4mm}
   \caption{Existing SEQ (Clip-then-Round) with a bitwidth schedule shows an unstable training behavior with sudden loss spikes as predicted by Fig.~\ref{fig:extreme}. The new RelaxSEQ with Round-the-Clip has a smooth transition in loss with changing bitwidth. Although, the final loss after the end of schedule was similar for both Clip-then-Round vs. Round-then-Clip, we found that RelaxSEQ resulted in better benchmark scores (see Table~\ref{tab:mmluNewSeqOldSeq}).}
   \label{fig:lossExSEQNewSEQ}
\end{figure}

We noticed this issue early on during the development of RelaxQAT. Specifically, we show the unstable behavior of existing SEQ (Clip-then-Round) in Fig.~\ref{fig:lossExSEQNewSEQ} for an early train-
\begin{wraptable}{r}{5.5cm}
    \scalebox{1.0}{
    \addtolength{\tabcolsep}{-2pt}
    \fontsize{8.0pt}{5.75pt}\selectfont
    \begin{tabular}{lc}
    \toprule
    \textbf{Method} & \textbf{MMLU} \\
    \midrule
    ParetoQ SEQ + bit schedule & 62.47 \\
    \midrule
    \textbf{RelaxSEQ + bit schedule} & \textbf{62.96 {\color{ForestGreen}(+0.5\%)}} \\
    
    \bottomrule
    \end{tabular}\vspace{-5mm}
    }
    \caption{Clip-then-Round (Existing SEQ) vs. Round-then-Clip (RelaxSEQ): Early Benchmarking on MMLU.} 
    \label{tab:mmluNewSeqOldSeq}
\end{wraptable} 
ing run for the Qwen3-4B-Instruct-2507 model. In contrast, RelaxSEQ has a smoothly changing loss. Interestingly, we noticed that although Clip-then-Round shows loss spikes and unstable training, the final loss did recover and perform similarly to RelaxSEQ. However, we found that Clip-then-Round resulted in poor benchmark scores as shown in Table~\ref{tab:mmluNewSeqOldSeq}. Consequently, we chose RelaxSEQ (Eq.~\eqref{eq:relax_seq}) as our main QDQ equation in this work.

\begin{figure*}[t]
  \centering
   \includegraphics[width=1\linewidth]{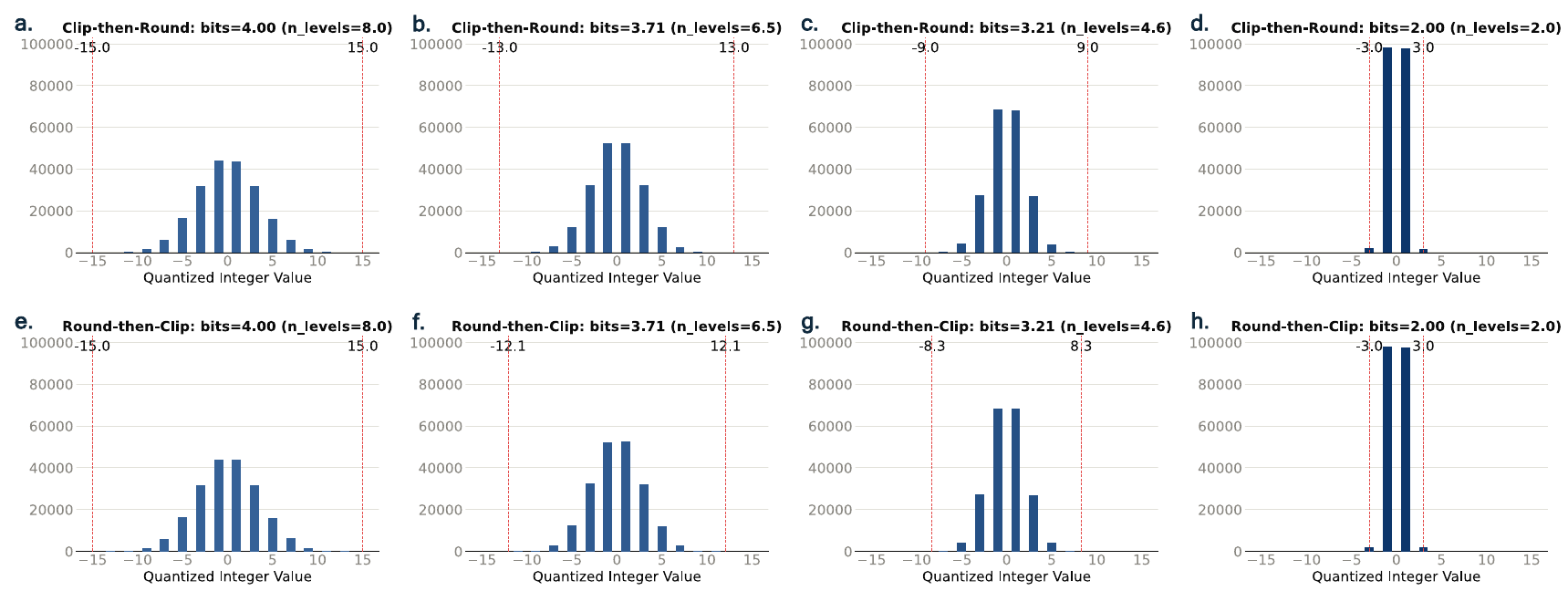}\vspace{-5mm}
   \caption{(a-d)~Clip-then-Round operation in the existing SEQ (Eq.~\eqref{eq:orig_seq}) creates integers at the extremes (see vertical red dotted lines) since the round operation follows the clip. As the bitwidth reduces, the extreme integers are \textit{suddenly} lost (e.g., from $\{-15, 15\}$ to $\{-13, 13\}$ and so on). This means that a training assumes that the previous range of integers (say $\{-15, 15\}$) was available until $k=2^{b-1}$ reduces sufficiently; once $k$ is sufficiently small, the round abruptly moves on to the new range of integers (i.e., $\{-13, 13\}$). The sudden change in integer ranges can cause loss spikes (similar to UPQ~\cite{upq}) and unstable training behavior. (e-h)~RelaxSEQ employs the Round-then-Clip operation (Eq.~\eqref{eq:relax_seq}) which keeps the extremes in floating point as the $k$ reduces. The range of the distribution changes slowly and the extremes stay floating point until the end of the schedule. This causes the loss to change smoothly as the bitwidth shrinks.}
   \label{fig:ctrrtc}
\end{figure*}

Figs.~\ref{fig:ctrrtc}(a-d) continue the analysis from Section~\ref{sec::relaxseq}. Using the same simulation environment from Fig.~\ref{fig:extreme}, we show how the weight distribution changes with bitwidth for the existing SEQ equation.  As $k=2^{b-1}$ shrinks with decreasing bitwidth, the weight distribution also narrows. However, the extreme values do not change until the argument to the round operation becomes sufficiently small. When $k$ becomes sufficiently small, the extreme integers are \textit{suddenly} lost. For instance, in Figs.~\ref{fig:ctrrtc}(a-d), the extreme integer values go from $\{-15, 15\}$ to $\{-13,-13\}$ and so on (see dotted red lines). This happens because in the existing SEQ equation, even if the bitwidth can be varied and is floating point, the extremes are still integers due to the Clip-then-Round operation. 

Figs.~\ref{fig:ctrrtc}(e-h) illustrate the intermediate floating point extreme values for RelaxSEQ. The extreme values for floating point bitwidths are floating points, yet all other intermediate quantization bins are integers.

\begin{figure*}[t]
  \centering
   \includegraphics[width=1\linewidth]{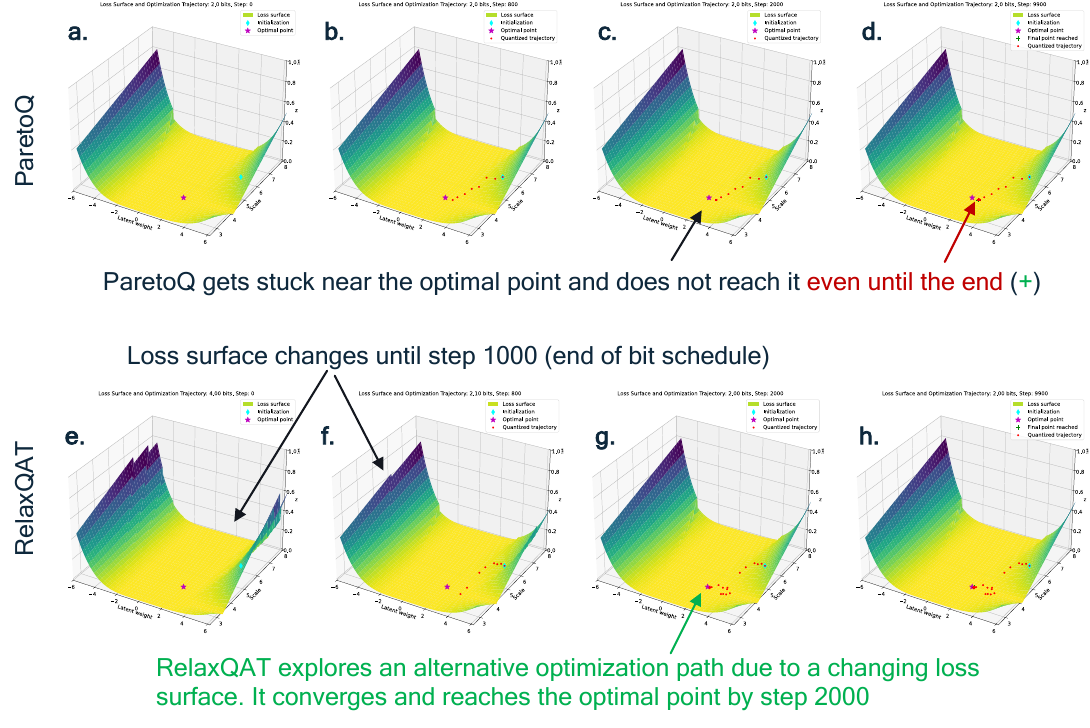}\vspace{-7mm}
   \caption{(a)~ParetoQ SEQ simulation for a 1-d optimization problem. QAT gets stuck near the minimum point around step 2000 and stays there until the end of training (10000 steps). (b)~RelaxQAT (with just a half cycle bit schedule) converges within the first 2000 steps.}
   \label{fig:losssim}
\end{figure*}

\section{Proof of Proposition 1}\label{app:prop1proof}
\setcounter{prop}{0}
\begin{prop}
The higher the bit precision, the smaller the width of the quantization hyperplanes. Specifically, given two bitwidths $b_1$ and $b_2$, with $b_1 < b_2$, the width of the quantization hyperplanes can be reduced by a factor of $2^{b_2-b_1}$ if we relax the bitwidth from $b_1$ to $b_2$. 

Therefore, relaxing the bitwidth during QAT can help the optimization escape the flat regions in the loss surface.
\label{prop1}
\end{prop}
\begin{proof}
Consider the LSQ Eq.~\eqref{eq:lsq} and let $\delta=\frac{\alpha}{k}=\frac{\alpha}{2^{b-1}}$. Then, the argument to the round operation becomes $\frac{\theta}{\delta}$. Note that, the round operation decides the boundaries of the quantization bins. Given a current integer value $z_i$, let us assume that the adjacent quantization bins lead to integers $z_{i-1}$ and $z_{i+1}$. Suppose that $\frac{\theta}{\delta}$ is at the center of the quantization bin for the integer $z_i$. Then, the minimum movement in $\frac{\theta}{\delta}$ to reach the neighboring quantization bins $z_{i-1}$ or $z_{i+1}$ would be $0.5$. For example, if $z_{i-1}=-1$, $z_i=0$, and $z_{i+1}=+1$, and $\frac{\theta}{\delta} = 0$, we need to move it by $0.5$ to reach the neighboring quantization boundaries. This is because $\frac{\theta}{\delta}\in (-0.5, 0.5)$ still results in $z_i=0$, $\frac{\theta}{\delta}=-0.5$ leads to $z_{i-1}=-1$, and $\frac{\theta}{\delta}=0.5$ leads to $z_{i+1}=+1$.

Therefore, the latent weight $\theta$ must move by $0.5\delta$ from the center of the quantization bin to change the integer value in the resulting QDQ weight. Hence, the width of the quantization hyperplane (defined by the flat region where the integer value does not change) is given by $=2\times 0.5\delta=\delta$. 

Now, since $\delta=\frac{\alpha}{2^{b-1}}$, the higher the bit precision, the smaller the delta, which in turn results in a smaller width of the quantization hyperplanes. Specifically, if we relax the bitwidth from $b_1$ to $b_2$ bits ($b_1<b_2$), we get:
\begin{equation}
    \frac{\delta_1}{\delta_2} = \frac{\frac{\alpha}{2^{b_1-1}}}{\frac{\alpha}{2^{b_2-1}}} = 2^{b_2-b_1}\implies \delta_2 = \frac{\delta_1}{2^{b_2-b_1}}
    \label{eq:prop1proof}
\end{equation}

That is, the width of the quantization hyperplanes reduces by a factor of $2^{b_2-b_1}$. As a result, if the QAT optimization is stuck in a flat region, dynamically and gradually relaxing the bitwidth can reduce the width of the quantization hyperplanes and enable the QAT to escape those flat regions.
\end{proof}

\section{SEQ Simulation}\label{app:seqsim}
We start again with a 1-d optimization problem which becomes 2-d for the quantized domain. For RelaxQAT, we also implement a bit schedule to see how a changing loss surface alters the optimization trajectory. For ParetoQ, the number of bits are fixed to 2-bits. Figs.~\ref{fig:losssim}(a-d) show that ParetoQ gets stuck near the true minimum and does not reach it even with many more (10000) training steps. In contrast, Fig.~\ref{fig:losssim}(e-h) converges within the 2000 steps because the initial changing loss surface sets the optimization on a different path. 

We notice similar convergence patterns on real LLM/VLM trainings (see Fig.~\ref{fig:tl} in the main paper) where (\textit{i})~the initial convergence accelerates significantly due to changing integers during the first half cycle of the bit schedule, and (\textit{ii})~Further $n$ cycles help the optimization escape the flatter regions in the loss surface. 

\section{More Details on Experiments}\label{app:moreinfo}
We use a total batch size of 256 for Qwen3-0.6B, Qwen3-4B and Gemma3-1B and a batch size of 320 for Qwen3.5-2B and Qwen3.5-4B models. We implement the AdamW optimizer with zero weight decay. 

For evaluation, we select a variety of tasks. For instance, common sense reasoning (CSR) benchmarks include five tasks: Arc-easy, Arc-Challenge, HellaSwag, PIQA and WinoGrande which focus on everyday knowledge tasks, grade-school science questions and plausibility. We also evaluate on benchmarks such as MMLU and IFEval. MMLU evaluates the model on factual recall and basic reasoning across 57 subjects whereas IFEval evaluates whether the model is following natural language instructions. While the CSR and MMLU tasks are multiple-choice questions, comparing the log-likelihood of each answer, IFEval is a generation task. We report accuracy norm metric for Arc-Easy, Arc-Challenge, Hellaswag, PIQA and accuracy for WingoGrande. IFEval reports four metrics namely prompt-level-strict-acc, inst-level-strict-acc, prompt-level-loose-acc, inst-level-loose-acc. For our experiments, we take an average of all four metrics.  

\section{Ablation Results}\label{app:ablations}
Gemma3-1B results are shown in Table~\ref{tab:g31bllm}.
\begin{table*}[t] 
    \centering 
    \scalebox{0.92}{ \setlength{\tabcolsep}{6pt} 
    \renewcommand{\arraystretch}{1.2} 
    \fontsize{8.0pt}{8.5pt}\selectfont 
    \begin{tabular}{l|c||ccccc||c|c|c||c} 
    \hline 
    \textbf{Method} & \textbf{Type} & \textbf{ArcE($\uparrow$)} & \textbf{ArcC($\uparrow$)} & \textbf{HS($\uparrow$)} & \textbf{WG($\uparrow$)} & \textbf{PIQA($\uparrow$)} & \textbf{CSR($\uparrow$)} & \textbf{MMLU($\uparrow$)} & \textbf{IFEval($\uparrow$)} & \textbf{Global Avg.($\uparrow$)} \\ 
    \hline 
    \noalign{\vskip 0.5pt} 
    \hline 
    FP & - & 63.43 & 38.31 & 57.79 & 58.48 & 72.42 & 58.09 & 41.37 & 73.38 & 57.61 \\ 
    \hline 
    AdaScale~\cite{adascaleptq} & PTQ & 28.49 & 23.89 & 27.72 & 49.56 & 53.48 & 36.63 & 23.75 & 9.80 & 23.39 \\ 
    \hline ParetoQ~\cite{paretoq} & \multirow{6}{*}{QAT} & 52.48 & \textbf{30.12} & 46.13 & \textbf{58.41} & 67.90 & \underline{51.01} & 25.28 & 53.12 & 43.14 {\color{ForestGreen}(+0.00\%)} \\ 
    UPQ~\cite{upq} & & 53.03 & 29.01 & 45.64 & 56.83 & \underline{68.17} & 50.54 & 25.34 & 42.01 & 39.30 {\color{red}(-3.84\%)} \\ LLM-QAT~\cite{llmqat} & & 25.97 & 25.68 & 26.14 & 50.43 & 51.63 & 35.97 & 22.95 & -$^*$ & -$^*$ \\ 
    BitNet~\cite{bitnet158hf} & & 52.82 & 28.92 & 45.24 & 56.99 & 67.63 & 50.32 & \underline{26.23} & 46.61 & 41.05 {\color{red}(-2.09\%)} \\ 
    WinQ~\cite{winq} & & \textbf{53.62} & 28.75 & \textbf{46.33} & \underline{58.17} & \textbf{68.88} & \textbf{51.15} & 25.64 & \underline{54.80} & \underline{43.86} {\color{ForestGreen}(+0.72\%)} \\ 
    \textbf{RelaxQAT (ours)} & & \underline{53.45} & \underline{29.10} & \underline{46.14} & 58.09 & 67.85 & 50.93 & \textbf{28.24} & \textbf{60.31} & \textbf{46.49} {\color{ForestGreen}\textbf{(+3.35\%)}} \\ \hline 
    \end{tabular} } 
    \vspace{-3mm} 
    \caption{Evaluation of Gemma3-1B on Commonsense Reasoning, MMLU, and IF-Eval benchmarks. WG and HS denote WinoGrande and HellaSwag, respectively. ($\uparrow$): higher is better. {\color{ForestGreen} Green} and {\color{red} red} denote improvement and degradation relative to ParetoQ, respectively. \textbf{Bold} is best and \underline{underline} is second-best among QAT methods. $^*$LLM-QAT produced repeating tokens for IFEval, hence, its accuracy cannot be reported.} \label{tab:g31bllm} 
\end{table*}

\section{Ternary Quantization Results}\label{app:ternaryquantization}
The ternary (1.58-bit) RelaxSEQ equation is given as:
\begin{equation}
    w_r = \frac{\alpha}{k} \left\{ \text{clip}\left(\text{round}\left[ \frac{\theta}{\alpha} \times k \right], -k+0.5, k-0.5 \right)\right\}.
    \label{eq:ternary_relax_seq}
\end{equation}
Here, at the end of the schedule, $k=2^{b_e-1}=2^{1.58-1}\approx1.5$. Hence, the QDQ weight $w_r$ achieves the ternary $\{-1, 0, 1\}$ integer grid. Of note, during training, we vary the bit schedule from $b_s=3$ bits to $b_e=1.58$ bits.

\begin{table*}[t] 
    \centering 
    \scalebox{0.97}{ \setlength{\tabcolsep}{6pt} 
    \renewcommand{\arraystretch}{1.2} 
    \fontsize{8.0pt}{8.5pt}\selectfont 
    \begin{tabular}{l||ccccc||c|c|c||c} 
    \hline 
    \textbf{Method} & \textbf{ArcE($\uparrow$)} & \textbf{ArcC($\uparrow$)} & \textbf{HS($\uparrow$)} & \textbf{WG($\uparrow$)} & \textbf{PIQA($\uparrow$)} & \textbf{CSR($\uparrow$)} & \textbf{MMLU($\uparrow$)} & \textbf{IFEval($\uparrow$)} & \textbf{Global Avg.($\uparrow$)} \\ 
    \hline 
    \noalign{\vskip 0.5pt} 
    \hline 
    FP & 81.22 & 58.44 & 69.08 & 67.48 & 75.73 & 70.39 & 72.63 & 86.55 & 76.52 \\ 
    \hline 
    ParetoQ~\cite{paretoq} & \textbf{75.08} & \textbf{47.44} & 58.28 & 61.56 & 71.87 & \underline{62.85} & 54.27 & 68.18 & 61.77 {\color{ForestGreen}(+0.00\%)} \\ 
    UPQ~\cite{upq} & 72.81 & 45.14 & \underline{58.48} & \underline{62.90} & \underline{72.52} & 62.37 & \underline{55.88} & \underline{74.88} & 64.38 {\color{ForestGreen}(+2.61\%)} \\
    BitNet~\cite{bitnet158hf} & 70.96 & 44.88 & 57.16 & 60.93 & 71.49 & 61.08 & 54.46 & 74.87 & 63.47{\color{ForestGreen}(+1.70\%)} \\ 
    WinQ~\cite{winq} & \underline{74.49} & 45.48 & 56.48 & 61.01 & 72.25 & 61.94 & 53.15 & 72.41 & 62.50 {\color{ForestGreen}(+0.73\%)} \\ 
    \textbf{RelaxQAT (ours)} & 73.23& \underline{47.35} & \textbf{60.89} & \textbf{65.35} & \textbf{73.50} & \textbf{64.06} & \textbf{58.96} & \textbf{74.90} & \textbf{65.97}{\color{ForestGreen}\textbf{(+4.20\%)}} \\ \hline 
    \end{tabular} } 
    \vspace{-3mm} 
    \caption{Evaluation of Qwen3-4B Ternary model on Commonsense Reasoning, MMLU, and IF-Eval benchmarks. WG and HS denote WinoGrande and HellaSwag, respectively. ($\uparrow$): higher is better. {\color{ForestGreen} Green} and {\color{red} red} denote improvement and degradation relative to ParetoQ, respectively. \textbf{Bold} is best and \underline{underline} is second-best among QAT methods.} \label{tab:tq34bllm} 
\end{table*}

\begin{table*}[t] 
    \centering 
    \scalebox{0.97}{ \setlength{\tabcolsep}{6pt} 
    \renewcommand{\arraystretch}{1.2} 
    \fontsize{8.0pt}{8.5pt}\selectfont 
    \begin{tabular}{l||ccccc||c|c|c||c} 
    \hline 
    \textbf{Method} & \textbf{ArcE($\uparrow$)} & \textbf{ArcC($\uparrow$)} & \textbf{HS($\uparrow$)} & \textbf{WG($\uparrow$)} & \textbf{PIQA($\uparrow$)} & \textbf{CSR($\uparrow$)} & \textbf{MMLU($\uparrow$)} & \textbf{IFEval($\uparrow$)} & \textbf{Global Avg.($\uparrow$)} \\ 
    \hline 
    \noalign{\vskip 0.5pt} 
    \hline 
    FP & 66.03 & 41.46 & 62.06 & 62.58 & 72.52 & 60.93 & 57.19 & 72.84 & 63.65 \\ 
    \hline 
    ParetoQ~\cite{paretoq} & \textbf{62.54} & \textbf{37.63} & 54.82 & \underline{61.40} & 70.24 & \textbf{57.33} & \underline{50.31} & 57.35 & \underline{55.00} {\color{ForestGreen}(+0.00\%)} \\ 
    UPQ~\cite{upq} & 28.66 & 22.35 & 24.65 & 51.38 & 50.38 & 35.49$^*$ & 24.08$^*$ & 18.34$^*$ & 25.97$^*$ {\color{red}(-29.03\%)} \\
    BitNet~\cite{bitnet158hf} & 56.73 & 35.24 & 54.09 & 60.06 & 69.48 & 55.12 & 47.31 & \underline{60.49} & 54.31 {\color{red}(-0.69\%)} \\ 
    WinQ~\cite{winq} & \underline{61.99} & \underline{36.86} & \underline{54.92} & \textbf{61.56} & \textbf{71.00} & \underline{57.27} & 50.28 & 56.60 & 54.72 {\color{red}(-0.28\%)} \\ 
    \textbf{RelaxQAT (ours)} & 59.93 & 36.69 & \textbf{55.33} & 61.25 & \underline{70.29} & 56.70 & \textbf{51.64} & \textbf{60.80} & \textbf{56.38} {\color{ForestGreen}\textbf{(+1.38\%)}} \\ \hline 
    \end{tabular} } 
    \vspace{-3mm} 
    \caption{Evaluation of Qwen3.5-2B Ternary model on Commonsense Reasoning, MMLU, and IF-Eval benchmarks. WG and HS denote WinoGrande and HellaSwag, respectively. ($\uparrow$): higher is better. {\color{ForestGreen} Green} and {\color{red} red} denote improvement and degradation relative to ParetoQ, respectively. \textbf{Bold} is best and \underline{underline} is second-best among QAT methods. $^*$UPQ did not converge well for the ternary training for Qwen3.5-2B, which explains its loss of accuracy.} \label{tab:tqwn352bllm} 
\end{table*}


\end{document}